\documentclass[runningheads]{llncs}
\usepackage{graphicx}

\usepackage{tikz}
\usepackage{comment}
\usepackage{amsmath,amssymb} 
\usepackage{color}

\usepackage[accsupp]{axessibility}  

\usepackage{xspace}
\usepackage{colortbl}
\usepackage{enumitem}
\usepackage{multirow}
\usepackage{makecell}
\usepackage{threeparttable}
\usepackage[export]{adjustbox}
\usepackage{graphics}
\usepackage{graphbox}
\usepackage{rotating}
\usepackage{bbm}

\usepackage{silence}
\usepackage[caption=false]{subfig}

\usepackage{tabularx}
\newcolumntype{Y}{>{\centering\arraybackslash}X}

\usetikzlibrary{spy, backgrounds}
\usetikzlibrary{positioning}

\definecolor{pending}{RGB}{51, 102, 255}
\definecolor{img-pending}{RGB}{204, 51, 255}
\definecolor{unfinish}{RGB}{0, 128, 0}
\definecolor{duplicate-text}{RGB}{255, 153, 0}
\definecolor{grey-text}{RGB}{128, 128, 128}

\newcommand*{\GENERALCOMPILE}{}  

\newcommand*{\eg}{e.g.\@\xspace}
\newcommand*{\ie}{i.e.\@\xspace}
\newcommand*{\etal}{et~al.\@\xspace}

\makeatletter
\newcommand*{\etc}{%
    \@ifnextchar{.}%
        {etc}%
        {etc.\@\xspace}%
}
\makeatother

\usepackage{orcidlink}
\usepackage{hyperref}

\begin{document}
\pagestyle{headings}
\mainmatter
\def\ECCVSubNumber{4988}  

\title{PD-Flow: A Point Cloud Denoising Framework with Normalizing Flows}


\titlerunning{PD-Flow}
%
\author{Aihua Mao\inst{1}\orcidlink{0000-0001-6861-9414} \and
Zihui Du\inst{1}\orcidlink{0000-0002-1096-2030} \and
Yu-Hui Wen\inst{2}\orcidlink{0000-0001-6195-9782} \and
Jun Xuan\inst{1}\orcidlink{0000-0002-7975-7852} \and
Yong-Jin Liu\inst{2}\orcidlink{0000-0001-5774-1916}
}
\authorrunning{A. Mao et al.}
%
\institute{South China University of Technology, Guangzhou, China
\email{ahmao@scut.edu.cn,\{csusami,202020143921\}@mail.scut.edu.cn}\\ \and
Tsinghua University, Beijing, China \\
\email{\{wenyh1616,liuyongjin\}@tsinghua.edu.cn}}
\maketitle

\renewcommand{\thefootnote}{} 

\footnotetext{Y.H Wen and Y.J Liu  are the corresponding authors.}
\begin{abstract}
Point cloud denoising aims to restore clean point clouds from raw observations corrupted by noise and outliers while preserving the fine-grained details.
We present a novel deep learning-based denoising model, that incorporates normalizing flows and noise disentanglement techniques to achieve high denoising accuracy. Unlike existing works that extract features of point clouds for point-wise correction, we formulate the denoising process from the perspective of distribution learning and feature disentanglement. By considering noisy point clouds as a joint distribution of clean points and noise, the denoised results can be derived from disentangling the noise counterpart from latent point representation, and the mapping between Euclidean and latent spaces is modeled by normalizing flows. We evaluate our method on synthesized 3D models and real-world datasets with various noise settings. Qualitative and quantitative results show that our method outperforms previous state-of-the-art deep learning-based approaches. The source code is available at \url{https://github.com/unknownue/pdflow}.
\end{abstract}

\setcounter{section}{0}

\section{Introduction}
\label{sec:intro}

As one of the most widely used representations for 3D objects, point clouds have attracted considerable attention in many fields, including geometric analysis, robotic object detection, and autonomous driving. The rapid development of 3D scanning devices, such as depth cameras and LiDAR sensors, has made point cloud data increasingly popular.
However, raw point clouds produced from these devices are inevitably contaminated by noise and outliers, due to inherent environment noise (\eg, lighting and background) and hardware limitation. Hence, point cloud denoising, which is a technique to restore high-quality and well-distributed points, is crucial for downstream tasks.

\begin{figure}[t]
\centering
\ifdefined\GENERALCOMPILE
    \ifdefined\HIGHRESOLUTIONFIGURE
        \includegraphics[width=\textwidth]{images/plotting/banner-high-resolution.pdf}
    \else
        \includegraphics[width=\textwidth]{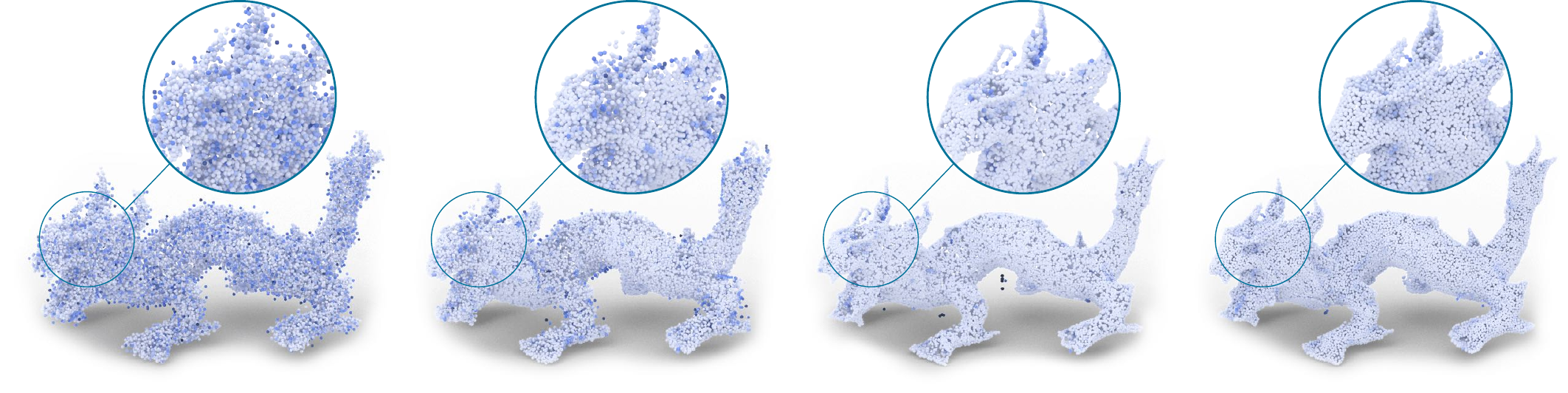}
    \fi
\fi
\begin{tabularx}{\textwidth}{@{}Y@{}Y@{}Y@{}Y@{}}
(a) Noisy data &
(b) DMRDenoise~\cite{luo2020differentiable} &
(c) ScoreDenoise~\cite{Luo_2021_ICCV} &
(d) Ours
\end{tabularx}
\caption{
The denoising results produced by (b) DMRDenoise~\cite{luo2020differentiable}, (c) ScoreDenoise~\cite{Luo_2021_ICCV} and (d) our method from noisy input (a).
Deeper color indicates higher error.
Our method preserves notably more fine details with less noise and outperforms others especially in uniformity.
}
\label{fig:banner-denoise-results}
\end{figure}

Despite decades of research, point cloud denoising remains a challenging problem, because of the intrinsic complexity of the topological relationship and connectivity among points.
Traditional denoising methods~\cite{alexa_pointset,lipman2007parameterization,avron2010,deschaud2010point,mattei2017point,lu2017gpf} perform well in some circumstances.
However, they generally rely on prior knowledge on point sets or some assumptions on noise distributions, and they may compromise the denoising quality for unseen noise (\eg, distortion, non-uniformity).

Recent promising deep learning approaches~\cite{duan20193d,rakotosaona2020pointcleannet,hermosilla2019totaldenoise,luo2020differentiable,Luo_2021_ICCV} bring new insight to point cloud denoising in a data-driven manner and exhibit superior performance over traditional methods.
These works can be classified into two categories.
The first class treats existing points as approximating the underlying surface by regressing points~\cite{duan20193d}, predicting displacements~\cite{rakotosaona2020pointcleannet,zhang2020pointfilter}, or progressive movement~\cite{Luo_2021_ICCV}.
Nonetheless, the point features are extracted from the local receptive field independently. Therefore, consistent surface properties may not be preserved between neighborhood points, resulting in artifacts, such as outliers and scatter. The second class treats downsampling noisy data as a coarse point set and resampling/upsampling points from the learned manifold with a target resolution~\cite{hermosilla2019totaldenoise,luo2020differentiable}.
However, the downsampling scheme inevitably discards geometric details, leading to distorted distribution.

In this paper, we consider the noisy point clouds as samples of the joint distribution of 3D shape and corrupted noise.
Based on this setup, it is intuitive to capture the characteristics of noise and underlying surface in the form of distribution.
Thus, we can formulate the point cloud denoising problem as disentangling the clean section from its latent representation. We can also interpret this idea from the perspective of signal processing~\cite{pauly2001spectral}, where clean points and noise are analogous to low- and high-frequency part of signals, respectively.
We can filter out the high-frequency contents and recover the smooth signal via the low-frequency counterpart that encodes the major information of raw signal.

\begin{figure}[t]
\centering
\includegraphics[width=\linewidth]{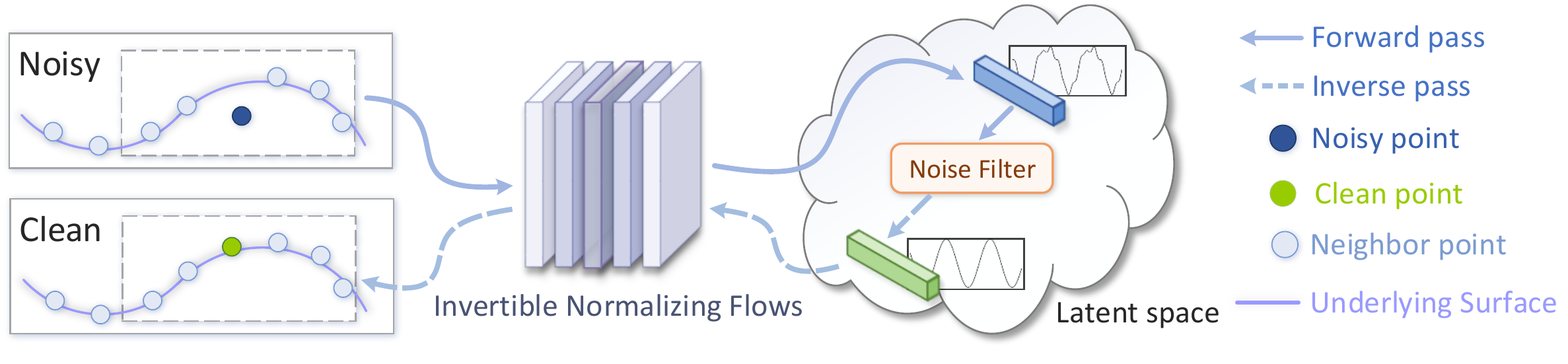}
\caption{
Schematic illustration of the proposed method.
We disentangle the noise factor from the latent representation of noisy point clouds and leverage NFs to model the transformation between Euclidean and latent spaces.
}
\label{fig:principle}
\end{figure}

Our denoising technique mainly consists of three phases: 1) learning the distribution of noisy point clouds by encoding the points into a latent representation, 2) filtering out the noise section from the latent representation, and 3) decoding/restoring noise-free points from the clean latent code.
To realize this process, we require a generative model that can simultaneously learn the latent distribution and restore clean points.
In this paper, we propose to exploit normalizing flows (NFs) in an invertible generative framework, to model the distribution mapping of point clouds. The whole process is illustrated in Fig.~\ref{fig:principle}.

Compared with other popular deep learning models, such as generative adversarial network (GAN) and variational autoencoder, NFs provide several advantages:
(\romannumeral1) NFs are capable of transforming complex distributions into disentangled code space, which is a desired property for point cloud denoising task,
(\romannumeral2) an NF is an invertible and lossless propagation process, which ensures one-to-one mapping between point clouds and their latent representations, and
(\romannumeral3) NFs realize the encoding and decoding process in a unified framework and share weights between forward and inverse propagations.

In summary, the main contributions of this work include:

\begin{itemize}
\item We propose a simple yet intuitive framework for point cloud denoising, called PD-Flow, which learns the distribution of noisy point sets and performs denoising via noise disentanglement.
\item We propose to augment vanilla flows to improve the flexibility and expressiveness of the latent representation of points.
We investigate various noise filtering strategies to disentangle noise from latent points.
\item To validate the effectiveness of our method, extensive evaluations are conducted on synthetic and real-world datasets.
Qualitative experiments show that our method outperforms the state-of-the-art works on diverse metrics.
\end{itemize}

\section{Related Works}

\subsection{Denoising Methods}
\noindent
{\bf Traditional denoising methods.}
Conventional methods for point cloud denoising can be coarsely classified into three categories:
1) Statistical-based filtering methods generally apply statistical analysis theories, such as kernel density estimation~\cite{schall2005robust}, sparse reconstruction principle~\cite{avron2010,sun2015denoising,mattei2017point}, principal component analysis~\cite{narvaez2006point}, Bayesian statistics~\cite{jenke2006bayesian} and curvature extraction~\cite{kalogerakis2009extracting}.
2) Projection-based filtering methods first construct a smooth surface (\eg, Moving Least Squares surface~\cite{alexa_pointset,fleishman2005robust,alexa2003computing}) from a set of noisy points. Then, denoising is implemented by projecting points onto surfaces.
According to projection strategies, this class of methods can be further divided into, e.g., locally optimal projection~\cite{lipman2007parameterization,huang2009consolidation,huang2013edge}, jet fitting~\cite{cazals2005estimating}, and bilateral filtering~\cite{digne2017bilateral}.
3) Neighborhood-based filtering methods measure the correlation and similarity between a point and its neighbor points.
Nonlocal-based methods~\cite{deschaud2010point,huhle2008robust,wang2008similarity,zheng2010non} generally detect self-similarity among nonlocal patches and consolidate them into coherent noise-free point clouds.
Graph-based denoising methods~\cite{gao2018graph,hu2020feature,schoenenberger2015graph,zeng20193d} naturally represent point cloud geometry with a graph. All above methods generally require user interaction or geometric priors (\eg, normals) and still lack the ability of filtering various noise levels.

\noindent
{\bf Deep-learning-based denoising methods.} In recent years, several deep learning based methods~\cite{rakotosaona2020pointcleannet,hermosilla2019totaldenoise,luo2020differentiable,duan20193d,Luo_2021_ICCV,pistilli2020learning} have been proposed for point cloud denoising. PointCleanNet~\cite{rakotosaona2020pointcleannet} first removes outliers and then predicts inverse displacement for each point~\cite{GuerreroEtAl_PCPNet}. It is the first learning-based method that directly inputs noisy data without the acquisition of normals nor noise/device speciﬁcations. Hermosilla \etal proposed Total Denoising (TotalDn)~\cite{hermosilla2019totaldenoise} to regress points from the distribution of unstructured \emph{total} noise.
This allows TotalDn to approximate the underlying surface without the supervision of clean points. Pistilli \etal proposed GPDNet~\cite{pistilli2020learning}, which is a graph convolutional network, to improve denoising robustness under high noise levels. In the denoising pipeline of DMRDenoise~\cite{luo2020differentiable}, noisy input is first downsampled by a differentiable pooling layer, and then the denoised points are resampled from estimated manifolds. However, using the downsampling schema~\cite{rakotosaona2020pointcleannet,luo2020differentiable} to remove outliers may cause unnecessary detail loss.
Recently, Luo and Hu developed a score-based denoising algorithm (ScoreDenoise~\cite{Luo_2021_ICCV}), which utilizes the gradient ascent technique and iteratively moves points to the underlying surface via estimated scores.

Our method differs from the above methods in several aspects.
First, we formulate the denoising process as disentangling noise from the factorized representation of noisy input.
Second, instead of applying separate modules to extract high-level features and reconstruct coordinates, we unify the point encoding/decoding process with a bijective network design.


\subsection{Normalizing Flows for Point Cloud Analysis}
NFs define a probability distribution transformation for data, allowing exact density evaluation and efficient sampling. In recent years, NFs have become a promising method for generative modeling and have been adopted into various applications~\cite{prenger2019waveglow,kumar2019videoflow,lugmayr2020srflow,abdal2021styleflow}. Representative models include discrete normalizing flows (DNF)~\cite{dinh2014nice,iclr_DinhSB17,kingma2018glow} and continuous normalizing flows (CNF)~\cite{chen2018neural,grathwohl2019ffjord}.

As the first NF-based algorithm for point cloud generation, PointFlow~\cite{yang2019pointflow} employs CNF to learn a two-level distribution hierarchy of given shapes.
PointFlow is a flexible scheme for modeling point distribution. However, the expensive equation solvers and training instability issues still remain to be open problems.
Sharing the similar idea, Discrete PointFlow~\cite{klokov20eccv} proposes to use discrete flow layers as an alternative to continuous flows to reduce computation overhead. Pumarola \etal~\cite{pumarola2020c} introduced C-Flow, which is a parallel conditional scheme in the DNF-based architecture, to bridge data between images and point clouds domains.
Postels \etal~\cite{postels2021go} recently presented mixtures of NFs to improve the representational ability of flows and show superior performance to a single NF model~\cite{klokov20eccv}.
These works mainly focus on improving generative ability and are evaluated on toy datasets. However, there are few works concentrating on flow-based real-world point cloud applications.

In this paper, we take advantage of the invertible capacity of NFs, which enable exact latent variable inference and efficient clean point synthesis.
To the best of our knowledge, no prior work has been proposed for the point cloud denoising task by developing a new framework with NFs.

\section{Method}

\subsection{Overview}

Given an input point set $\mathcal{\tilde{P}}=\left\{\tilde{p}_{i}=p_{i}+o_i\right\}\in\mathbb{R}^{N\times D_p}$ corrupted by noise $\mathcal {O}=\left\{o_i\right\}$, we aim to predict a clean point set $\mathcal{\hat{P}}=\left\{\hat p_{i}\right\}\in \mathbb{R}^{N\times D_p}$, where $N$ is the number of points, $D_p$ is the point coordinate dimension, and $\hat{p}_i$ is the prediction of clean point $p_i$.
In our study, we consider the coordinate dimension with $D_p=3$ and make no assumptions about the noise distribution of $\mathcal {O}$.

\begin{figure}[t]
\centering
\includegraphics[width=\linewidth]{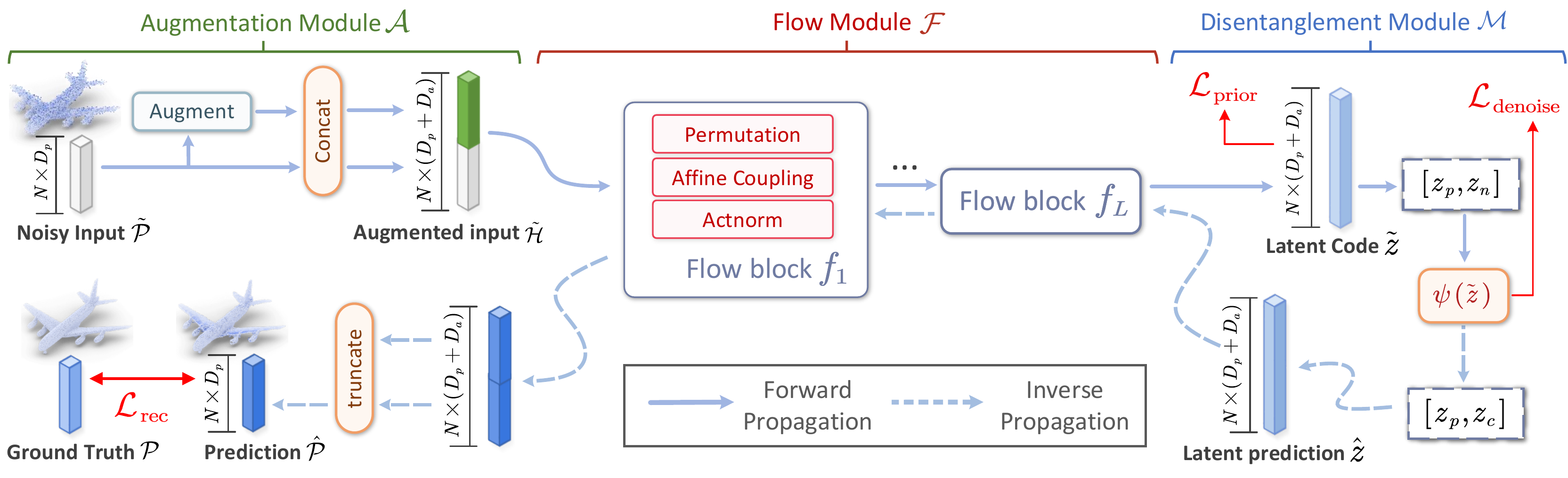}
\caption{
The proposed flow-based denoising framework.
Given a noisy point set $\mathcal{\tilde{P}}=\left\{\tilde{p}_{i}\in\mathbb{R}^{D_p}\right\}$, we first augment it with an additional $D_a$ dimensional variable $\bar{p}_i\in\mathbb{R}^{D_a}$ and obtain augmented point $h_i=\left[\tilde{p}_i, \bar{p}_i\right]\in\mathbb{R}^{D_p+D_a}$.
We transform the augmented points $\mathcal{\tilde{H}}=\left\{ h_{i}\right\}$ to latent distribution $\tilde{z}\sim p_{\vartheta}(\tilde{z})$ by NFs. To estimate noise-free latent point $\hat{z}$, we filter out the noise factor from noisy $\tilde{z}$. We restore the noise-free point set $\mathcal{\hat{P}}$ from $\hat{z}$ by the inverse propagation of $\mathcal F$, which utilizes the invertible capacity of NFs.
Finally, the coordinates of the clean point set are derived from truncating the first $D_p$ dimensions and discarding the augmented $D_a$ dimensions.
}
\label{fig:network-arch}
\end{figure}

In this paper, we propose to utilize NFs to model the mapping of point distribution between Euclidean and latent spaces, thereby allowing us to formulate point cloud denoising as the problem of disentangling the noise factor from its latent representation. The overall denoising pipeline is shown in Fig.~\ref{fig:network-arch}.


\subsection{Flow-based Denoising Method}
\label{sec:flow-denoising-background}

We consider the point cloud denoising problem from the perspective of distribution learning and disentanglement.
We suppose the distribution of noisy point set $\mathcal{\tilde{P}}$ is the joint distribution of clean point set $\mathcal{P}=\left\{p_i\right\}$ and noise $\mathcal{O}$.
Given a dataset of observation $\mathcal{\tilde{P}}$, we aim to learn a bijective mapping $f_{\theta}$, which is parameterized by $\theta$ to approximate the data distribution:
\begin{equation}
\label{eq:noise_p_encode}
\tilde{z}=f_{\theta}(\mathcal{\tilde{P}})=f_{\theta}(\mathcal{P},\mathcal{O}), 
\end{equation}
where $\tilde{z}\sim p_{\vartheta}(\tilde{z})$ is a random variable with known probability density. Note that $p_{\vartheta}(\tilde{z})$ follows a factorized distribution~\cite{dinh2014nice}, such that $p_{\vartheta}(\tilde{z})=\prod_{i} p_{\vartheta}\left(\tilde{z}_{i}\right)$ (\ie the dimensions of $\tilde{z}$ are independent of each other).

We further assume that $f_{\theta}$ can simultaneously learn to embed noise factor and intrinsic structure of point cloud into a disentangled latent code space (\ie where noise is uniquely controlled by some dimensions).
Based on this assumption, we approximate the clean latent representation $z$ by
\begin{equation}
\label{eq:noise_z_disentanglement}
\hat{z}=\psi\left(\tilde{z}\right),
\end{equation}
where $\psi:\mathbb{R}^D\rightarrow\mathbb{R}^D$ is a disentanglement function defined in latent space, and $\hat{z}$ is an estimation of $z$. In this way, clean point samples $\mathcal{\hat{P}}$ can be derived by taking the inverse transformation
\begin{equation}
\label{eq:clean_p_decode}
\mathcal{\hat{P}}=g_\theta\left(\hat{z}\right),
\end{equation}
where $g_{\theta}(\cdot)=f_{\theta}^{-1}(\cdot)$.
The bijective mapping $f_{\theta}$, which consists of a sequence of invertible transformations $f^1_{\theta},\cdots,f^L_{\theta}$, is referred to as \emph{normalizing flows}.
Denote by $h^l$ the output of $l$-th flow transformation.
Then $h^{l+1}$ can be formulated as
\begin{equation}
h^{l+1}=f_{\theta}^{l+1}\left(h^{l} \right),
\end{equation}
where $h^0=\mathcal{\tilde{P}}$, $h^{L}=\tilde{z}$.
Applying the change-of-variables formula and chain rule~\cite{dinh2014nice}, the output probability density of $\mathcal{\tilde{P}}$ can be obtained as
\begin{equation}
\label{eq:log_pdf_noise_P}
\begin{aligned}
\log p(\mathcal{\tilde{P}}; \theta)&=\log p_{\vartheta}\left(f_{\theta}(\mathcal{\tilde{P}})\right)+\log \left|\operatorname{det} \frac{\partial f_{\theta}}{\partial \mathcal{\tilde{P}}}(\mathcal{\tilde{P}})\right| \\&=\log p_{\vartheta}\left(f_{\theta}(\mathcal{\tilde{P}})\right)+\sum_{l=1}^{L} \log \left|\operatorname{det} \frac{\partial f_{\theta}^{l}}{\partial h^{l}}\left(h^{l}\right)\right|,
\end{aligned}
\end{equation}
where $\left|\operatorname{det} \frac{\partial f_{\theta}}{\partial \mathcal{\tilde{P}}}(\mathcal{\tilde{P}})\right|$ is the log-absolute-determinant of the Jacobian of mapping $f_{\theta}$, which measures the volume change~\cite{dinh2014nice} caused by $f_{\theta}$.
$f_{\theta}$ can be trained with the maximum likelihood principle using the gradient descent technique.

\subsection{Augmentation Module}
\label{sec:method_augment_module}
\begin{figure}[t]
\centering
\subfloat[Traditional flow transformation\label{fig:flow-traditional}]{
\includegraphics[width=0.45\linewidth]{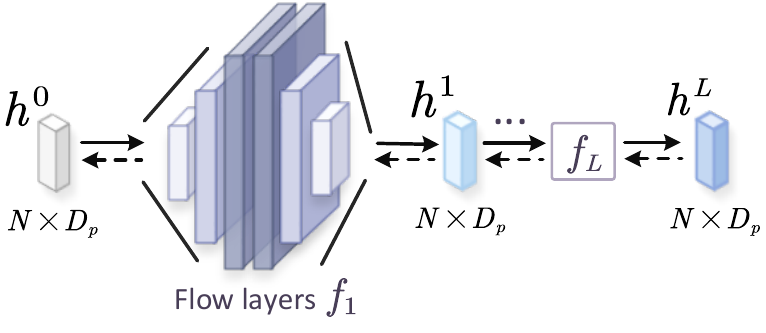}
}
\subfloat[Augmented flow transformation\label{fig:flow-augmented}]{
\includegraphics[width=0.50\linewidth]{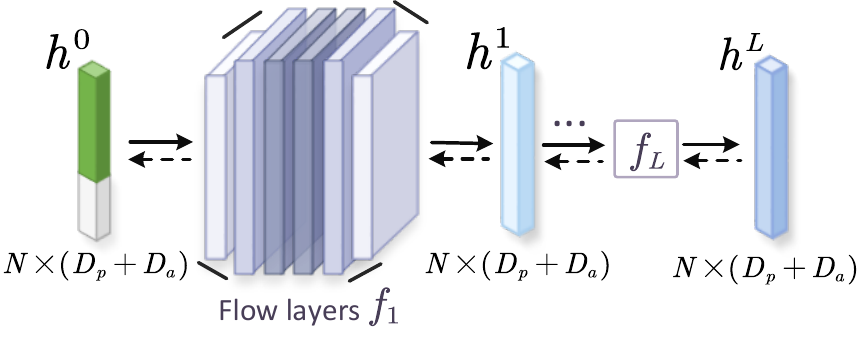}
}
\caption{
Illustration of dimension augmentation.
(a) The dimensionality of raw data limits the bandwidth of the network in vanilla flows.
(b) Augmented flows can share considerable information among flow blocks, thereby improving model expressiveness.
}
\label{fig:augmentation}
\end{figure}

\noindent
{\bf Dimensional bottleneck.} To maintain the analytical invertibility, flow models impose more constraints on the network architecture than non-invertible models. One particular constraint is that the flow components $f^1,\cdots,f^L$ must output the same dimensionality $D$ with the input data (where $D=3$ for raw point clouds). The network bandwidth bottleneck sacrifices the model expressiveness, as shown in Fig.~\ref{fig:flow-traditional}.

Previous works~\cite{iclr_DinhSB17,kingma2018glow} generally use a squeezing operator to alleviate this limitation by exchanging spatial dimensions for feature channels.
However, the squeezing operator is mainly designed for image manipulation. It is non-trivial to adopt squeezing to point cloud due to the unorder nature in point sets.

\noindent
{\bf Dimension augmentation.} Inspired by VFlow~\cite{chen2020vflow}, we resolve the bottleneck by increasing the dimensionality of input data.
To be specific, for each input point $\tilde{p}_{i}\in\mathbb{R}^{D_p}$ in $\mathcal{\tilde{P}}$, we augment it with a random variable $\bar{p}_i\in\mathbb{R}^{D_a}$.
This process is modeled by an augmentation module $\mathcal A$:
\begin{equation}
\bar{p}_i=\mathcal{A}\left(\tilde{p_i}, \mathcal{N}\left(\tilde{p_i}\right)\right), \mathcal{\bar{P}}=\left\{\bar{p}_i\right\},
\end{equation}
where $\mathcal{N}\left(\tilde{p_i}\right)$ denotes the $k$-nearest neighbors of $\tilde{p_i}$, and $\mathcal{\bar{P}}$ represents the set of augmented dimensions. 
We feed the augmented point set $\mathcal{\tilde{H}}=\left\{ h_{i}=\left[\tilde{p}_i, \bar{p}_i\right]\right\}$ as input of flow module $\mathcal F$, and the underlying NFs become $\tilde{z}=f_{\theta}(\mathcal{\tilde{H}})$, where $\tilde{z}\in\mathbb{R}^{D_p+D_a}$, as shown in Fig.~\ref{fig:flow-augmented}.

\noindent
{\bf Variational augmentation.} To model the distribution of the augmented data space, VFlow~\cite{chen2020vflow} resorts to optimizing the evidence lower bound observation (ELBO) on the log-likelihood of augmented data as an alternative of Eq.~\ref{eq:log_pdf_noise_P}:
\begin{equation}
\label{eq:vflow_elbo}
\log p(\mathcal{\tilde{P}}; \theta)\geq\mathbb{E}_{q(\mathcal{\bar{P}}\mid \mathcal{\tilde{P}};\phi)}\left[\log p(\mathcal{\tilde{P}},\mathcal{\bar{P}};\theta)-\log q(\mathcal{\bar{P}}\mid \mathcal{\tilde{P}};\phi)\right],
\end{equation}
where $q(\mathcal{\bar{P}}\mid \mathcal{\tilde{P}};\phi)$ indicates the distribution of augmented data, which is modeled by the augmentation module $\mathcal A$, $\theta$ and $\phi$ denote the parameters of $\mathcal{F}$ and $\mathcal{A}$, respectively. We briefly explain Eq.~(\ref{eq:vflow_elbo}) in supplementary material.

\subsection{Flow Module}
\label{sec:method_flow_module}

The flow module $\mathcal F$ transforms augmented points $\mathcal{\tilde{H}}=\left\{ h_{i}\right\}\in \mathbb{R}^{N\times \left(D_p+D_a\right)}$ from the Euclidean space to the latent space, and vice versa.

The architecture of flow module $\mathcal F$ comprises $L$ blocks, where each block consists of a couple of flow components, as shown in Fig.~\ref{fig:network-arch}.
Each component is designed to satisfy the efficient invertibility and tractable Jacobian, including affine coupling layer~\cite{iclr_DinhSB17}, actnorm~\cite{kingma2018glow}, and permutation layer~\cite{kingma2018glow}. The descriptions of each flow component are detailed in supplementary material.

\subsection{Disentanglement Module}
\label{sec:method_disentanglement_module}

\begin{figure}[t]
\centering
\includegraphics[width=1.0\linewidth]{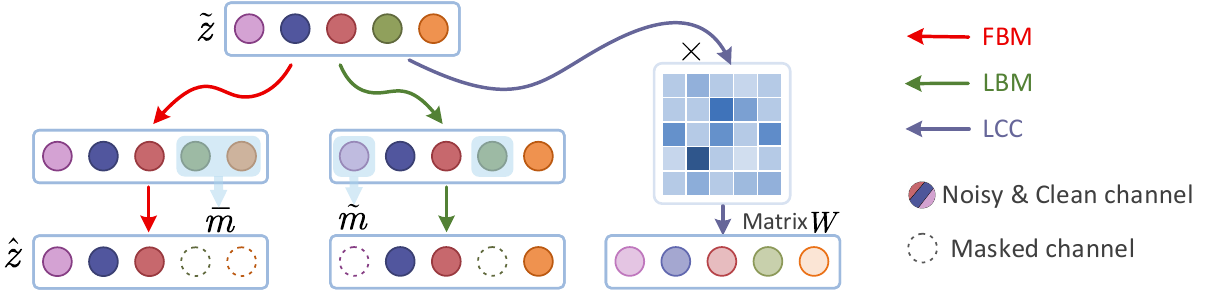}
\caption{Illustration of different filtering strategies.}
\label{fig:disentagle-methods}
\end{figure}

Let $z_p$ and $z_n$ be the clean point and noisy parts of the latent point $\tilde{z}$, \ie $\tilde{z}=[z_p, z_n]$.
We aim to disentangle noise $z_n$ from $\tilde{z}$ by a smooth operator $\psi:\mathbb{R}^D\rightarrow\mathbb{R}^D$
\begin{equation}
\hat{z}=[z_p, z_c]=\psi(\tilde{z}),
\end{equation}
where $z_c$ is the denoised feature and $D=D_p+D_a$.
$\hat{z}$ denotes the prediction of noise-free point representation, which is fed as the input of inverse propagation of flow module $\mathcal{F}$.
However, how to implement $\psi(\cdot)$ is non-trivial.
In this paper, we investigate three types of noise filtering strategies (Fig.~\ref{fig:disentagle-methods}) to formulate $\psi(\cdot)$ as follows:

\noindent
{\bf Fix Binary Mask (FBM).} Similiar to a previous work~\cite{liu2021disentangling}, we explicitly divides the channels of latent code into two groups, \ie clean and noisy channels.
FBM simply sets noisy channels to $0$ by
\begin{equation}
\label{eq:FBM-formulation}
\psi(\tilde{z})=\bar{m}\odot \tilde{z},
\end{equation}
\begin{equation}
\label{eq:FBM-loss}
\mathcal{L}_{\text{denoise}}=\mathcal{L}_{\text{FBM}}=0,
\end{equation}
where $\bar{m}\in\{0, 1\}^D$ is a fixed binary mask specified by the user, $\odot$ denotes element-wise product and $\mathcal{L}_{\text{FBM}}$ is the corresponding loss function of FBM.

\noindent
{\bf Learnable Binary Mask (LBM).} We employ a soft masking to latent $\tilde{z}$ by
\begin{equation}
\label{eq:LBM-formulation}
\psi(\tilde{z})=\tilde{m}\odot \tilde{z},
\end{equation}
\begin{equation}
\label{eq:LBM-loss}
\mathcal{L}_{\text{denoise}}=\mathcal{L}_{\text{LBM}}=\left|\tilde{m}\left(1-\tilde{m}\right)\right|,
\end{equation}
where $\tilde{m}\in\mathbb{R}^D$ is a learnable parameter, $|\cdot|$ denotes $L_1$ norm, and $\mathcal{L}_{\text{LBM}}$ is the corresponding loss of LBM that encourages $\tilde{m}$ to approximate the binary mask.

\noindent
{\bf Latent Code Consistency (LCC).} We minimize the latent representation between clean points and noisy points by
\begin{equation}
\label{eq:LCC-formulation}
\psi(\tilde{z})=W\tilde{z},
\end{equation}
\begin{equation}
\label{eq:LCC-loss}
\mathcal{L}_{\text{denoise}}=\mathcal{L}_{\text{LCC}}=\sum_{i=1}^{N}\|W\tilde{z}^{(i)}-z^{(i)}\|,
\end{equation}
where $W\in\mathbb{R}^{D\times D}$ is a learnable matrix to transform $\tilde{z}$, $N$ is the number of points, $\|\cdot\|$ denotes $L_2$ norm, and $z=f_{\theta}(\mathcal P)$ is the latent representation encoded from reference points $\mathcal{P}$ by the forward propagation of flow (similar to Eq.~(\ref{eq:noise_p_encode})).
$\mathcal{L}_{\text{LCC}}$ is the corresponding loss of LCC that encourages the transformed $\hat{z}$ to be consistent with noise-free representation $z$, which is analogous to perceptual loss ~\cite{Johnson2016Perceptual} that evaluates difference on high-level features.

\subsection{Joint Loss function}

We present an objective function for training PD-Flow that combines the \emph{reconstruction loss}, \emph{prior loss} and \emph{denoise loss} (Section \ref{sec:method_disentanglement_module}) as follows:

\noindent
{\bf Reconstruction loss} quantifies the similarity between the generated points $\mathcal{\hat{P}}\in\mathbb{R}^{N\times D_p}$ and reference clean points $\mathcal{P}\in\mathbb{R}^{N\times D_p}$.
In this paper, we use the Earth Mover’s Distance (EMD) metric as $\mathcal{L}_{\text{rec}}$ by minimizing
\begin{equation}
\mathcal{L}_{\text{rec}}=\mathcal{L}_{\mathrm{EMD}}(\hat{\mathcal{P}}, \mathcal{P})=\min _{\varphi: \hat{\mathcal{P}} \rightarrow \mathcal{P}} \sum_{\hat{p}\in \hat{\mathcal{P}}}\left\|\hat{p}-\varphi\left(\hat{p}\right)\right\|,
\end{equation}
where $\varphi: \hat{\mathcal{P}} \rightarrow \mathcal{P}$ is a bijection and $\left\|\cdot\right\|$ denotes $L_2$ norm.

\noindent
{\bf Prior loss} optimizes the transformation capability of flow module $\mathcal F$ by maximizing the likelihood of observation $\mathcal{\tilde{P}}$.
We implement the prior loss by minimizing the negative ELBO in Eq.~(\ref{eq:vflow_elbo}):
\begin{equation}
\label{eq:loss_prior}
\mathcal{L}_{\text{prior}}(\mathcal{\tilde{P}}) = \mathcal{L}(\mathcal{\tilde{P};\theta,\phi}) =-\left[\log p(\mathcal{\tilde{P}},\mathcal{\bar{P}};\theta)-\log q(\mathcal{\bar{P}}\mid \mathcal{\tilde{P}};\phi)\right]
\end{equation}
where $\mathcal{\bar{P}}=\mathcal{A}(\mathcal{\tilde{P}})$ are augmented dimensions (Section \ref{sec:method_augment_module}).
Intuitively, $\mathcal{L}_{\text{prior}}$ encourages the input points $\mathcal{\tilde{P}}$ to reach high probability under the predefined prior $p_{\vartheta}(\tilde{z})$.

\noindent
{\bf Total Loss.} Combining the preceding formulas, our method can be trained in an end-to-end manner by minimizing

\begin{equation}
\mathcal{L}\left(\theta,\phi,\sigma\right)= \alpha\mathcal{L}_{\text {rec}}+\beta\mathcal{L}_{\text {prior}}+\gamma\mathcal{L}_{\text{denoise}},
\label{eq:total_loss}
\end{equation}
where $\theta$, $\phi$ and $\sigma$ denotes the network parameters of $\mathcal F$, $\mathcal A$ and $\mathcal M$, respectively. And, $\alpha$, $\beta$, $\gamma$ are the hyper-parameters to balance the loss.

\subsection{Discussion}

\noindent
{\bf Benefit of dimension augmentation.} The dimension augmentation setting provides extra benefits to vanilla flows:
(\romannumeral1) The augmented NFs are generalization of vanilla flows, where the extra dimensionality $D_a$ can be freely adjusted by users, allowing it to model more complex function.
(\romannumeral2) The augmented dimensions afford more flexibility and expressiveness to intermediate point features (\ie $h^l$ in Section \ref{sec:flow-denoising-background}) between flow transformations, avoiding extracting high dimensional features from scratch.
(\romannumeral3) The augmented dimensions increase the degrees of freedom for noise filtering in the disentanglement phase, which is particularly helpful because 
raw point clouds contain only one dimension of $D_p=3$.
We investigate the influence of dimension augmentation in Section \ref{sec:ablation-study}.

Although the augmented dimensions increase the network size of flow module $\mathcal F$, the overhead is only marginal.
The computation overhead mainly depends on the hidden layer size $D_h$ of the internal transformation unit of $\mathcal F$ instead of the output dimensionality $D_p+D_a$.

\noindent
{\bf Unified noise disentanglement pipeline.} Considering the invertible property of NFs, raw points $\mathcal{\tilde{P}}$ and latent $\tilde{z}$ share the identical information in different domains.
We only manipulate the point features in the disentanglement module throughout the whole denoising pipeline, demonstrating the feature disentanglement capability of NFs.

Additionally, we do not explicitly introduce extra network modules to predict point-wise displacement~\cite{rakotosaona2020pointcleannet} or upsample to a target resolution~\cite{luo2020differentiable} for point generation.
Utilizing the flow invertibility can share parameters between forward and inverse propagations, which help us to reduce the network size and avoid the use of a decoding module.

\section{Experiments}

\subsection{Datasets}

We evaluate our method on the following datasets:
(\romannumeral1) {\bf PUSet.} This dataset is a subset of PUNet~\cite{yu2018pu} provided by \cite{Luo_2021_ICCV}, which contains 40 meshes for training and 20 meshes for evaluation.
(\romannumeral2) {\bf DMRSet.} This dataset collects meshes from ModelNet40~\cite{Zhirong15CVPR} provided by \cite{luo2020differentiable}, which contains 91 meshes for training and 60 meshes for evaluation. These point clouds are perturbed by Gaussian noise of various noise levels at resolutions ranging from 10K to 50K points.

We implement PD-Flow with the PyTorch framework. The training settings, datasets and network configurations are detailed in the supplementary material.

\begin{table}[t]
\centering
\caption{Comparison of denoising algorithms on PUSet.}
\resizebox{\linewidth}{!}{%
\setlength\tabcolsep{4pt}
\begin{tabular}{c| c c c | c c c | c c c || c c c | c c c | c c c }
\hline
\#Points & \multicolumn{9}{c||}{10K} & \multicolumn{9}{c}{50K} \\
\hline
Noise Level & \multicolumn{3}{c|}{1\%} & \multicolumn{3}{c|}{2\%} & \multicolumn{3}{c||}{3\%} & \multicolumn{3}{c|}{1\%} & \multicolumn{3}{c|}{2\%} & \multicolumn{3}{c}{3\%} \\
\hline
Method & \thead{CD \\ $10^{-4}$} & \thead{P2M \\ $10^{-4}$} & \thead{HD \\ $10^{-3}$} & \thead{CD \\ $10^{-4}$} & \thead{P2M \\ $10^{-4}$} & \thead{HD \\ $10^{-3}$} & \thead{CD \\ $10^{-4}$} & \thead{P2M \\ $10^{-4}$} & \thead{HD \\ $10^{-3}$} & \thead{CD \\ $10^{-4}$} & \thead{P2M \\ $10^{-4}$} & \thead{HD \\ $10^{-3}$} & \thead{CD \\ $10^{-4}$} & \thead{P2M \\ $10^{-4}$} & \thead{HD \\ $10^{-3}$} & \thead{CD \\ $10^{-4}$} & \thead{P2M \\ $10^{-4}$} & \thead{HD \\ $10^{-3}$} \\
\hline
Jet~\cite{cazals2005estimating} & 3.47 & 1.20 & 1.58 & 4.83 & 1.89 & 2.97 & 6.15 & 2.86 & 7.00 & 0.82 & 0.19 & 0.90 & 2.38 & 1.35 & 3.36 & 5.64 & 4.16 & 9.05 \\
GPF~\cite{lu2017gpf} & 3.28 & 1.17 & 1.52 & 4.18 & 1.54 & 3.45 & 5.37 & 2.75 & 8.14 & 0.76 & 0.23 & 1.42 & 2.04 & 0.94 & 4.25 & 3.82 & 2.87 & 10.4 \\
MRPCA~\cite{mattei2017point} & 3.14 & 1.01 & 1.77 & 3.87 & 1.26 & {\bf 2.59} & 5.13 & {\bf 2.03} & {\bf 4.84} & 0.70 & {\bf 0.12} & {\bf 0.79} & 2.11 & 1.06 & 3.21 & 5.64 & 3.97 & 7.65 \\
GLR~\cite{zeng20193d} & 2.79 & 0.92 & {\bf 1.16} & 3.66 & 1.14 & {\bf 2.88} & 4.84 & 2.08 & 6.80 & 0.71 & 0.18 & 0.93 & 1.61 & 0.85 & 4.90 & 3.74 & 2.67 & 11.7 \\
\hline
PCNet~\cite{rakotosaona2020pointcleannet} & 3.57 & 1.15 & 1.54 & 7.54 & 3.92 & 6.25 & 13.0 & 8.92 & 13.9 & 0.95 & 0.27 & 2.21 & 1.56 & 0.62 & 9.84 & 2.32 & 1.32 & 8.42 \\
GPDNet~\cite{pistilli2020learning} & 3.75 & 1.33 & 3.03 & 8.00 & 4.50 & 6.08 & 13.4 & 9.33 & 13.5 & 1.97 & 1.09 & 1.94 & 5.08 & 3.84 & 7.56 & 9.65 & 8.14 & 16.7 \\
Pointfilter~\cite{zhang2020pointfilter} & 2.86 & 0.75 & 2.87 & 3.97 & 1.30 & 6.21 & 4.94 & 2.14 & 9.26 & 0.82 & 0.24 & 2.38 & 1.46 & 0.77 & 4.58 & 2.25 & 1.44 & 8.70 \\
DMR~\cite{luo2020differentiable} & 4.54 & 1.70 & 6.72 & 5.04 & 2.13 & 7.02 & 5.87 & 2.86 & 8.60 & 1.17 & 0.46 & 2.26 & 1.58 & 0.81 & 4.29 & 2.45 & 1.54 & 7.32 \\
Score~\cite{Luo_2021_ICCV} & 2.52 & 0.46 & 4.30 & 3.68 & {\bf 1.08} & 5.78 & 4.69 & {\bf 1.94} & 10.5 & 0.71 & {\bf 0.15} & 2.30 & {\bf 1.28} & {\bf 0.57} & 4.95 & {\bf 1.92} & {\bf 1.05} & 9.30  \\
\hline
Ours & {\bf 2.12} & {\bf 0.38} & {\bf 1.36} & {\bf 3.25} & {\bf 1.02} & 3.71 & {\bf 4.45} & {\bf 2.05} & {\bf 5.31} & {\bf 0.65} & {\bf 0.16} & {\bf 1.71} & {\bf 1.18} & {\bf 0.60} & {\bf 2.58} & {\bf 1.94} & \bf 1.26 & {\bf 5.21} \\
\hline
\end{tabular}
}
\label{tab:comparison_sota_score}
\end{table}%

\subsection{Comparisons with State-of-the-art Methods}
\label{sec:experiment-comparison-sota}

\noindent
{\bf Evaluation metrics.}
We use four evaluation metrics in quantitative comparison, including (\romannumeral1) Chamfer distance (CD), (\romannumeral2) Point-to-mesh (P2M) distance, (\romannumeral3) Hausdorff distance (HD), and (\romannumeral4) Uniformity (Uni).
The detailed description of each metric is available in the supplementary material.

\noindent
{\bf Quantitative comparison.}
We compare our method with traditional methods (including Jet~\cite{cazals2005estimating}, MRPCA~\cite{mattei2017point}, GPF~\cite{lu2017gpf}, GLR~\cite{zeng20193d}) and state-of-the-art deep learning-based methods (including PointCleanNet (PCNet)~\cite{rakotosaona2020pointcleannet}, Pointfilter~\cite{zhang2020pointfilter}, DMRDenoise (DMR)~\cite{luo2020differentiable}, GPDNet~\cite{pistilli2020learning}, ScoreDenoise (Score)~\cite{Luo_2021_ICCV}).

We use LCC as the default noise filter.
The benchmark is based on 10K and 50K points disturbed by isotropic Gaussian noise with the standard deviation of noise ranging from 1\% to 3\% of the shape’s bounding sphere radius.

As shown in Table~\ref{tab:comparison_sota_score}, traditional methods can achieve good performance in some cases depending on the manual tuning parameters, but they have difficulty in extending to all metrics.
PCNet~\cite{rakotosaona2020pointcleannet} and DMRDenoise~\cite{luo2020differentiable} perform less satisfactory under 10K points, while GPDNet~\cite{pistilli2020learning} fails to handle high noise levels.
In most cases, our method outperforms Pointfilter~\cite{zhang2020pointfilter} and ScoreDenoise~\cite{Luo_2021_ICCV}, especially in CD and HD metrics.
The quantitative comparison on DMRSet and more results of various noise types are provided in the supplementary material.

\begin{table}[t]
\centering
\caption{
Comparison of uniformity on 10K points under various Gaussian noise levels.
This metric is estimated in the local area of different radii $p$.
Besides, we also show the corresponding CD loss ($\times10^{-4}$) of the full point clouds.
}
\setlength{\tabcolsep}{7pt}
\begin{tabular}{c|c|c|ccccc}
\hline
\multirow{2}{*}{Noise} & \multirow{2}{*}{Methods} & \multirow{2}{*}{CD} & \multicolumn{5}{c}{Uniformity for different $p$} \\
\cline{4-8}
& & & 0.4\% & 0.6\% & 0.8\% & 1.0\% & 1.2\% \\
\hline
\multirow{4}{*}{1\%} & MRPCA~\cite{mattei2017point} & 3.14 & 1.89 & 2.30 & 2.42 & 2.59 & 2.83 \\
& DMR~\cite{luo2020differentiable} & 4.54 & 4.02 & 5.06 & 6.02 & 7.03 & 7.95 \\
& Score~\cite{Luo_2021_ICCV} & 2.52 & 1.10 & 1.38 & 1.69 & 2.05 & 2.45 \\
& Ours & {\bf 2.12} & {\bf 0.33} & {\bf 0.43} & {\bf 0.55} &  {\bf 0.71} & {\bf 0.89} \\
\hline
\multirow{4}{*}{2\%} & MRPCA~\cite{mattei2017point} & 3.87 & 2.21 & 2.56 & 2.85 & 2.97 & 3.14 \\
& DMR~\cite{luo2020differentiable} & 5.04 & 3.45 & 4.04 & 4.68 & 5.35 & 6.03 \\
& Score~\cite{Luo_2021_ICCV} & 3.68 & 1.95 & 2.39 & 2.91 & 3.44 & 4.04 \\
& Ours & {\bf 3.25} & {\bf 0.89} & {\bf 1.18} & {\bf 1.49} &  {\bf 1.83} & {\bf 2.18} \\
\hline
\multirow{4}{*}{3\%} & MRPCA~\cite{mattei2017point} & 5.13 & 2.28 & {\bf 2.29} & {\bf 2.32} & {\bf 2.45} & {\bf 2.60} \\
& DMR~\cite{luo2020differentiable} & 5.87 & 3.81 & 4.53 & 5.16 & 5.85 & 6.55 \\
& Score~\cite{Luo_2021_ICCV} & 4.69 & 4.19 & 5.42 & 6.43 & 7.46 & 8.37 \\
& Ours & {\bf 4.45} & {\bf 1.80} & {\bf 2.33} & 2.83 & 3.34 & 3.87 \\
\hline
\end{tabular}
\label{tab:comparison_uniformity}
\end{table}%

\begin{figure}[t]

\ifdefined\GENERALCOMPILE
    \null\hfill
    \begin{tikzpicture}
    \node (fig) at (current page.east) {\includegraphics[width=0.15\textwidth]{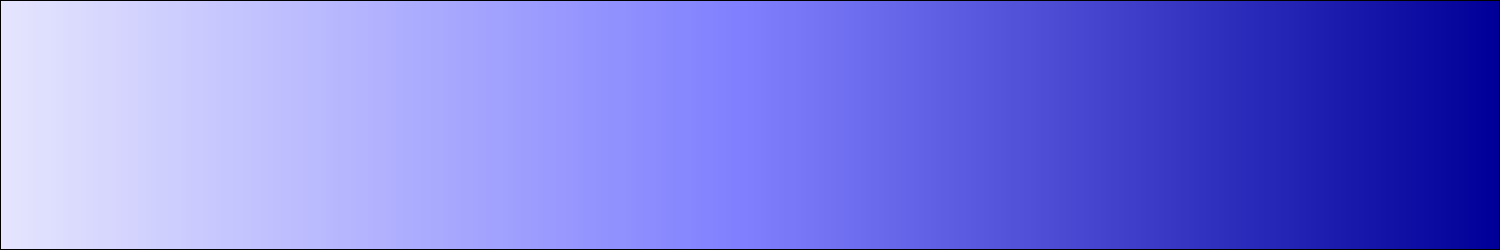}};
    \node[left=0cm of fig] {Clean};
    \node[right=0cm of fig] {Noisy};
    \end{tikzpicture}

    \centering
    \ifdefined\HIGHRESOLUTIONFIGURE
        \includegraphics[width=0.98\textwidth]{images/plotting/sota_comparison-high-resolution.pdf}
    \else
        \includegraphics[width=0.98\textwidth]{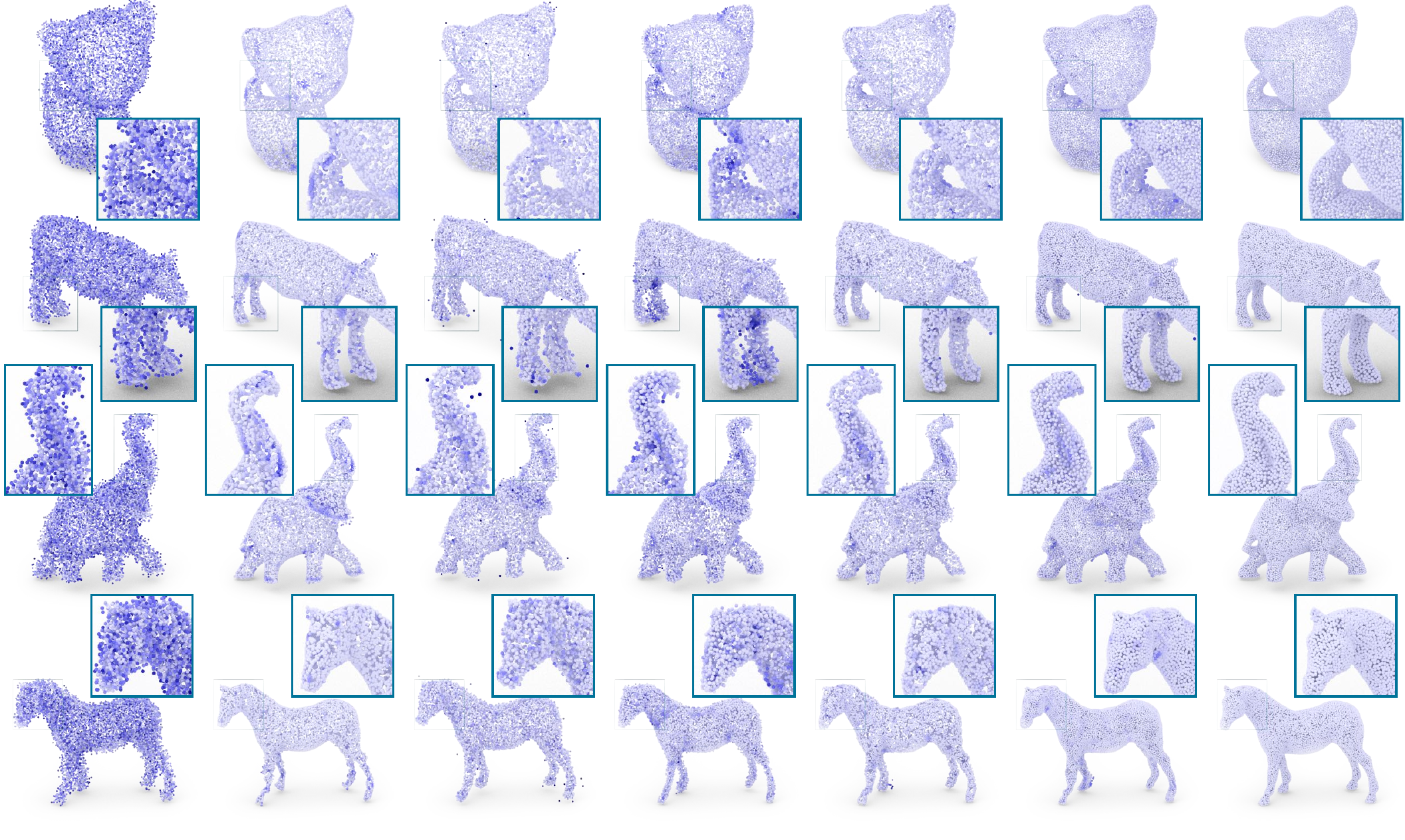}
    \fi
\fi
\centering
\begin{tabularx}{0.98\textwidth}{@{}Y@{}Y@{}Y@{}Y@{}Y@{}Y@{}Y@{}}
\scalebox{0.85}[0.85]{(a) Noisy} &
\scalebox{0.75}[0.75]{(b)MRPCA\cite{mattei2017point}} &
\scalebox{0.85}[0.85]{(c) PCNet\cite{rakotosaona2020pointcleannet}} &
\scalebox{0.85}[0.85]{(d) DMR \cite{luo2020differentiable}} &
\scalebox{0.85}[0.85]{(e) Score \cite{Luo_2021_ICCV}} &
\scalebox{0.85}[0.85]{(f) Ours} &
\scalebox{0.85}[0.85]{(g) Clean}
\end{tabularx}
\caption{
Visual comparison of state-of-the-art denoising methods under 2\% isotropic Gaussian noise.
The color of each point indicates its reconstruction error measured by P2M distance.
See supplementary material for visual results under more noise settings.
}
\label{fig:visual-comparison-sota}
\end{figure}
\begin{figure}[t]
\centering
\ifdefined\GENERALCOMPILE
    \ifdefined\HIGHRESOLUTIONFIGURE
        \includegraphics[width=0.93\textwidth]{images/plotting/RueMadame/RueMadame6-high-resolution.pdf}
    \else
        \includegraphics[width=0.93\textwidth]{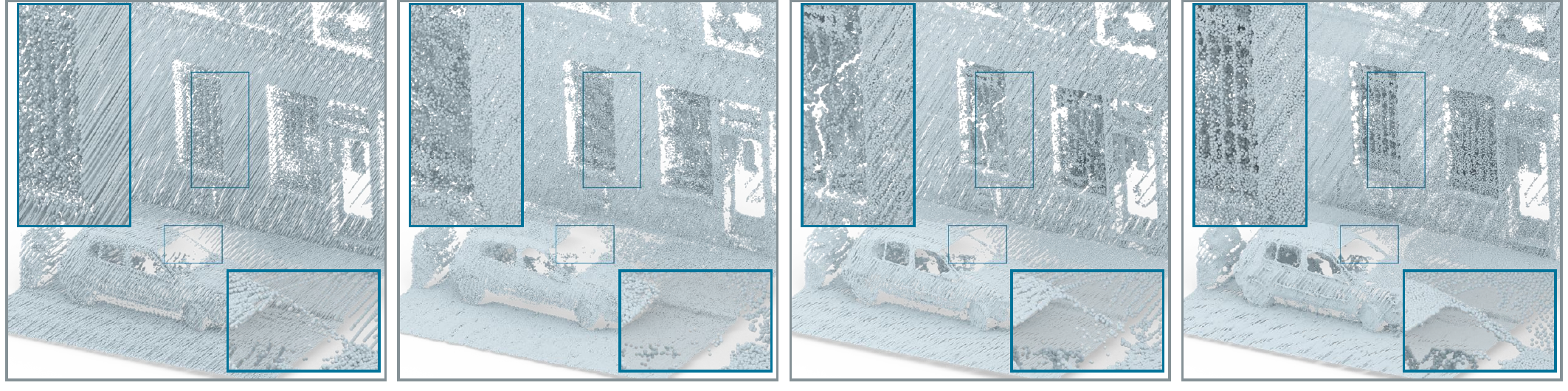}
    \fi
\fi

\begin{tabularx}{0.93\textwidth}{@{}Y@{}Y@{}Y@{}Y@{}@{}}
\scalebox{0.93}[0.93]{(a) Noisy} &
\scalebox{0.93}[0.93]{(b) DMRDenoise~\cite{luo2020differentiable}} &
\scalebox{0.93}[0.93]{(c) ScoreDenoise~\cite{Luo_2021_ICCV}} &
\scalebox{0.93}[0.93]{(d) Ours}
\end{tabularx}

\caption{Visual results of our denoiser on the real-world dataset \emph{Paris-rue-Madame}~\cite{serna2014paris}.}
\label{fig:visual-comparison-RueMadame}
\end{figure}

Furthermore, we present the quantitative results on uniformity metric under various noise levels in Table~\ref{tab:comparison_uniformity}.
Although MRPCA~\cite{mattei2017point} achieves the best uniformity under 3\% Gaussian noise, it fails to keep good generation accuracy on CD metric.
Compared with other state-of-the-art methods~\cite{luo2020differentiable,Luo_2021_ICCV}, our method considerably promotes the uniformity of generated points.

\noindent
{\bf Qualitative comparison.}
Fig.~\ref{fig:visual-comparison-sota} visualizes the qualitative denoising results between ours and competitive works.
We observe that our method achieves the most robust estimation under high noise corruption.
In particular, our method can keep consistent density across different regions and avoid clustering phenomenon, resulting in better uniformity.

We also compare the denoising result under the \emph{Paris-rue-Madame}~\cite{serna2014paris} dataset, which contains real-world scene data captured by laser scanner.
As shown in Fig.~\ref{fig:visual-comparison-RueMadame}, our method improves the surface smoothness and preserves better details than DMRDenoise~\cite{luo2020differentiable} and ScoreDenoise~\cite{Luo_2021_ICCV}.

\noindent
\subsection{Ablation Study}
\label{sec:ablation-study}

We conduct ablation studies to demonstrate the contribution of the network design of PD-Flow.
The evaluation is based on 10K points with 2\% Gaussian noise in PUSet.

\noindent
{\bf Flow architecture.} The number of parameters mainly depends on the depth of flow module $\mathcal{F}$ (\eg, the number of flow blocks $L$).
As shown in Table~\ref{tab:ablation_flow_architectures}, the fitting capacity improve as $L$ increases.
However, when $L$ increases to 12, the relative performance boost becomes marginal, with the cost of a large number of network parameters and training instability.
We find that $L=8$ achieves the best balance between performance and training stability.

\begin{figure}[t]
\centering
\subfloat[Ablation on augmentation\label{fig:ablation_augment_channels}]{
\includegraphics[width=0.42\columnwidth]{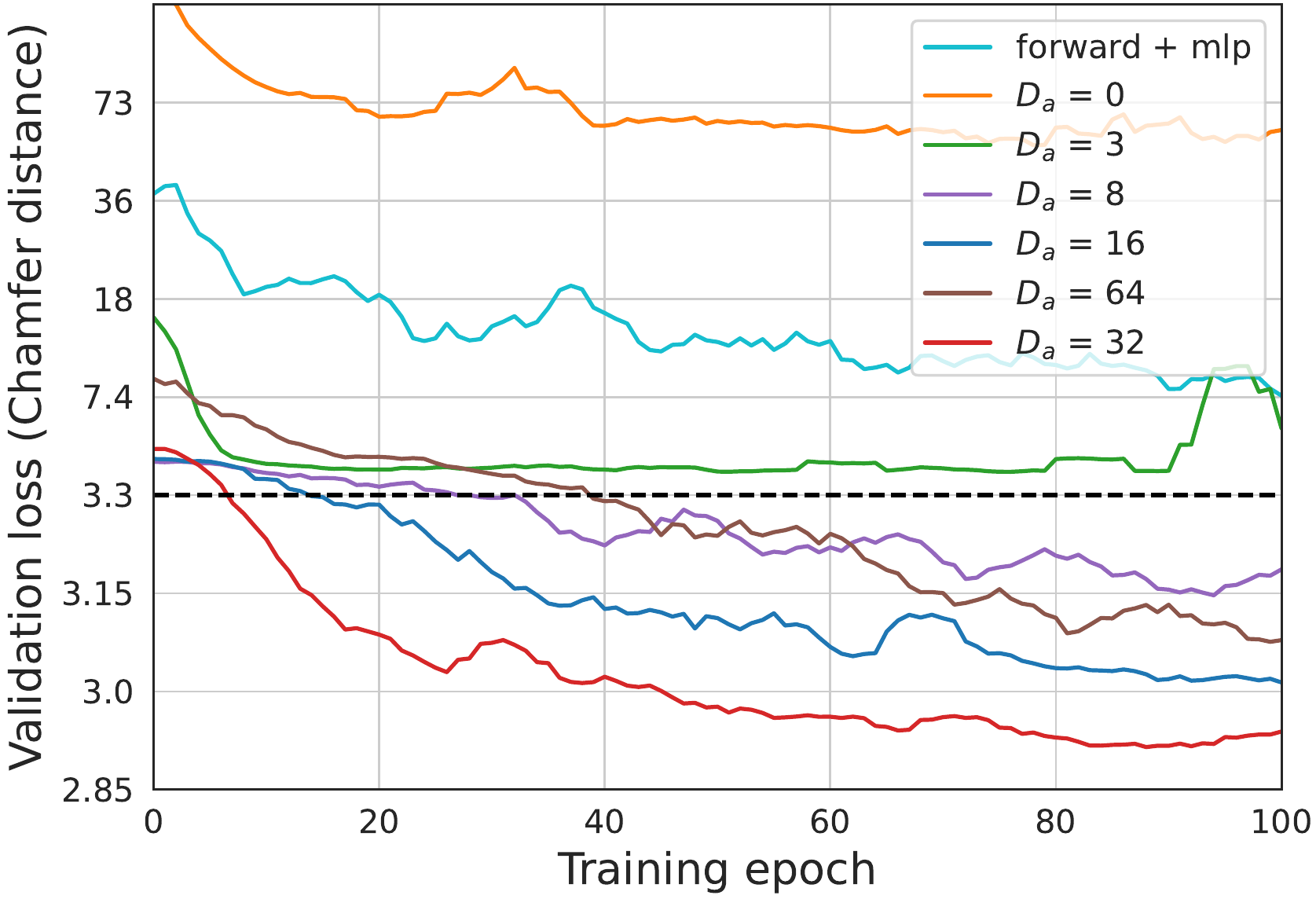}
}
\subfloat[Ablation on FBM filter\label{fig:ablation_masked_channels}]{
\includegraphics[width=0.42\columnwidth]{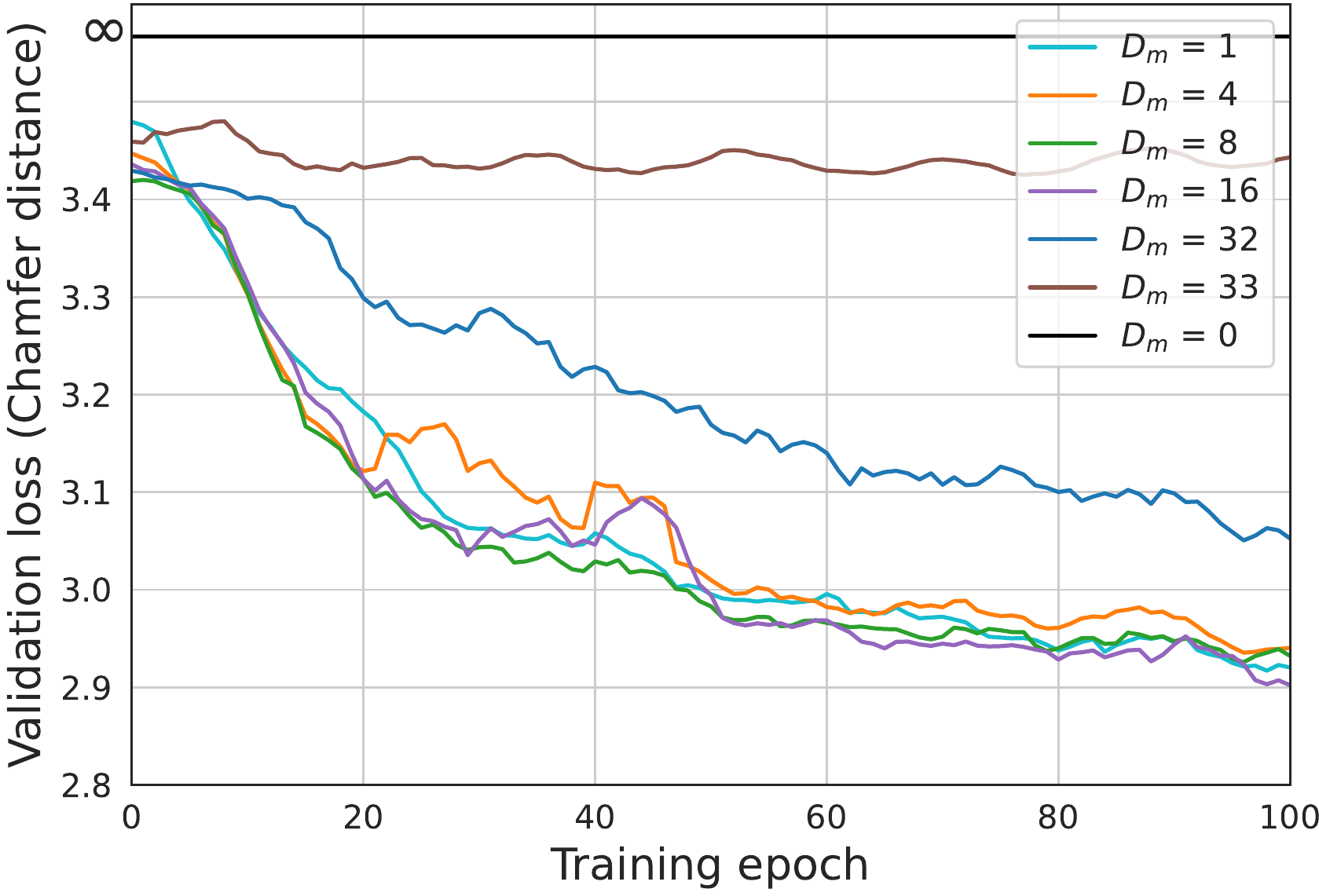}
}
\caption{
(a) Loss curves of training PD-Flow with augmentation channels 
and LCC noise filter.
For better visualization, we use linear mapping below dashed line and non-linear mapping above dashed line.
(b) Loss curves of training PD-Flow with FBM noise filter.
All these models are augmented with $D_a=32$ and vary in mask channels $D_m$.
}
\end{figure}

To verify the effectiveness of inverse propagation of $\mathcal{F}$, we replace the inverse pass with MLP layers, which are commonly used in other deep-learning-based methods~\cite{rakotosaona2020pointcleannet,luo2020differentiable}.
As shown in Table~\ref{tab:ablation_flow_architectures}, the inverse pass achieves better generation quality without introducing extra network parameters, which demonstrates the feasibility of flow invertibility.
The similar result is also verified by the ``forward + mlp" curve in Fig.~\ref{fig:ablation_augment_channels}, where using MLP as point generator leads to degraded performance.

\begin{table}[t]
\centering
\caption{Ablation study of flow architectures.}
\setlength{\tabcolsep}{7pt}
\begin{tabular}{l|c|c|c|c}
\hline
\multicolumn{1}{c|}{Pass} & \#Flow block & \#Params & CD & P2M \\ \hline
Forward + Inverse & 4  & 299K & 3.55 & 1.22 \\ \hline
Forward + Inverse & 8  & 470K & {\bf 3.25} & {\bf 1.03} \\ \hline
Forward + Inverse & 12 & 647K & 3.57 & 1.23 \\ \hline
Forward + MLP     & 8  & 578K & 4.65 & 2.14 \\ \hline
\end{tabular}
\label{tab:ablation_flow_architectures}
\end{table}

\noindent
{\bf Dimension augmentation.} To investigate the impact of the number of augmentation channels $D_a$ on model convergence, we show the training curve in Fig.~\ref{fig:ablation_augment_channels}.
The baseline model with $D_a=0$ (\ie vanilla flows) fails to converge to reasonable results.
As the $D_a$ increases, we observe faster convergence and better fitting capability.
The similar trending can also be observed in quantitative evaluation under different $D_a$.
This indicates that the dimension augmentation makes a key contribution to activate the denoising capability of NFs.

\noindent
{\bf Effect on noise filtering strategies.} We compare the performances of various filtering strategies with $D_a = 32$.
As shown in Table~\ref{tab:ablation_filtering_strategies}, all these strategies can achieve competitive performance, and the LCC filter achieves the best results.
For LBM with $D_a=16$ to $D_a=64$, we observe that about 2/3 channels approximate zeros in $\tilde{m}$ after training.

We further investigate the impact of the number of masked channels $D_m$ on FBM filter in Fig.~\ref{fig:ablation_masked_channels}, which shows the training curve under different $D_m$ settings.
For $D_m=0$, the flow network produces the same result as input due to the invertible nature of NFs.
For $D_m=1$ to $D_m=16$, our method can converge to reasonable performance and presents little differences in convergence.
This indicates that our method can adaptively embed noisy and clean channels to disentangled position, even if $D_m$ is set to a low value.
For $D_a=32$ and $D_m\geq 32$, the latent code space only contains 3 or less channels to embed intrinsic information of clean points.
We can observe that the insufficient channels for clean point embedding obviously lead to degrade performance, demonstrating that the dimension of latent point has great impact on model expressiveness and denoising flexibility. In Section \ref{sec:method_disentanglement_module}, we introduce $\mathcal{L}_{\text{denoise}}$ as a regularization term to improve noise disentanglment capability.
Table~\ref{tab:ablation_filtering_strategies} compares the effects of $\mathcal{L}_{\text{denoise}}$ term.
The ``LBM (w/o $\mathcal{L}_{\text{LBM}}$)" term means that the experiment is trained without the $\mathcal{L}_{\text{LCC}}$ loss (Eq.~(\ref{eq:LBM-loss})), but still uses LBM as the filter strategy (Eq.~(\ref{eq:LBM-formulation})).
As shown in Table~\ref{tab:ablation_filtering_strategies}, both $\mathcal{L}_{\text{FBM}}$ and $\mathcal{L}_{\text{LCC}}$ loss make positive contributions to model performance.

\begin{table}[t]
\centering
\caption{
Ablation study of different filtering strategies.
}
\setlength{\tabcolsep}{7pt}
\begin{tabular}{l|cc|cc|cc}
\hline
\#Points, Noise & \multicolumn{2}{c|}{10K,1\%} & \multicolumn{2}{c|}{10K,2\%} & \multicolumn{2}{c}{10K,3\%} \\
\hline
\multicolumn{1}{c|}{Strategy} & CD & P2M & CD & P2M & CD & P2M \\ \hline
FBM                              & 2.22 & 0.44 & 3.40 & 1.14 & 4.56 & 2.16 \\ \hline
LBM (w/o $\mathcal{L}_{\text{LBM}}$) & 2.54 & 0.62 & 3.65 & 1.38 & 5.06 & 2.32 \\ \hline
LBM (w/ $\mathcal{L}_{\text{LBM}}$) & 2.37 & 0.53 & 3.50 & 1.23 & 4.80 & 2.23 \\ \hline
LCC (w/o $\mathcal{L}_{\text{LCC}}$) & 2.32 & 0.45 & 3.48 & 1.19 & 4.62 & 2.20 \\ \hline
LCC (w/ $\mathcal{L}_{\text{LCC}}$) & {\bf 2.13} & {\bf 0.38} & {\bf 3.26} & {\bf 1.03} & {\bf 4.50} & {\bf 2.09} \\ \hline
\end{tabular}
\label{tab:ablation_filtering_strategies}
\end{table}

\section{Conclusion}

In this paper, we present PD-Flow, a point cloud denoising framework that combines NFs and distribution disentanglement techniques. It learns to transform noise perturbation and clean points into a disentangled latent code space by leveraging NFs, whereas denoising is formulated as channel masking. To alleviate the dimensional bandwidth bottleneck and improve the network expressiveness, we propose to extend additional channels to latent variables by an dimension augmentation module.
Extensive experiments and ablation studies illustrate that our method outperforms existing state-of-the-art methods in terms of generation quality across various noise levels.

\noindent
{\bf Acknowledgements:}
This work was supported by the NSF of Guangdong Province (2019A1515010833, 2022A1515011573), the Natural Science Foundation of China (61725204) and Tsinghua University Initiative Scientific Research Program, China Postdoctoral Science Foundation (2021M701891).



\clearpage
%
%
\bibliographystyle{splncs04}
\bibliography{egbib}

\newpage

\titlerunning{PD-Flow}
\authorrunning{A. Mao et al.}

\title{Supplementary Material of \\PD-Flow: A Point Cloud Denoising Framework with Normalizing Flows}
\author{}
\institute{}
\maketitle
\def\thesection{\Alph{section}}

\setcounter{equation}{17}
\setcounter{figure}{8}
\setcounter{table}{4}
\setcounter{page}{18}
\setcounter{section}{0}

\section{Estimating ELBO of Augmented Flows}

In this section, we present a brief description of the variational augmentation used in Section \ref{sec:method_augment_module}.
Please refer to VFlow~\cite{chen2020vflow} for the detailed theoretical proof.

The augmented data distribution $q(\mathcal{\bar{P}}\mid \mathcal{\tilde{P}};\phi)$ is modeled with a conditional flow
\begin{equation}
\mathcal{\bar{P}}=f^{-1}_{\phi}(\epsilon;\mathcal{\tilde{P}}),
\end{equation}
where $\epsilon\sim p_{\vartheta}(\epsilon)$ is a known prior distribution defined by user and is similar to $p_{\vartheta}(\tilde{z})$, $f^{-1}_{\phi}$ denotes the inverse propagation pass of flow $f_{\phi}$ (similar to Eq.~\ref{eq:clean_p_decode} in the main paper), and $\phi$ denotes the network parameters of augmentation module $\mathcal{A}$.
Here, $\mathcal{\tilde{P}}$ is used as the conditional input for $f^{-1}_{\phi}$ that helps generate the augmented dimensions $\mathcal{\bar{P}}$.
The probability density of $\mathcal{\bar{P}}$ can be computed as
\begin{equation}
\log q(\mathcal{\bar{P}}\mid\mathcal{\tilde{P}} ; \phi)=\log p_{\vartheta}\left(\epsilon\right)-\log \left|\frac{\partial \mathcal{\bar{P}}}{\partial \epsilon}\right|,
\end{equation}
where $\log q(\mathcal{\bar{P}}\mid\mathcal{\tilde{P}} ; \phi)$ is the second term in Eq.~\ref{eq:vflow_elbo} and Eq.~\ref{eq:loss_prior} in the main paper.
Please refer to Section \ref{sec:supplement-network-augment-module} for the detailed implementation of $f^{-1}_{\phi}$.

Learning the joint distribution of $\mathcal{\tilde{P}}$ and $\mathcal{\bar{P}}$ can be regarded as modeling $p(\mathcal{\tilde{H}}; \theta)$, where $\mathcal{\tilde{H}}=\left\{ h_{i}=\left[\tilde{p}_i, \bar{p}_i\right]\right\}$.
Thus, we can formulate the probability density of $\mathcal{\tilde{H}}$ by
\begin{equation}
\log p(\mathcal{\tilde{P}},\mathcal{\bar{P}};\theta) = \log p(\mathcal{\tilde{H}};\theta)=\log p_{\vartheta}\left(f_{\theta}(\mathcal{\tilde{H}})\right)+\log \left|\operatorname{det} \frac{\partial f_{\theta}}{\partial \mathcal{\tilde{H}}}(\mathcal{\tilde{H}})\right|,
\end{equation}
where $\theta$ denotes the network parameters of flow module $\mathcal F$. $\log p(\mathcal{\tilde{P}},\mathcal{\bar{P}};\theta)$ is the first term in Eq.~\ref{eq:vflow_elbo} and Eq.~\ref{eq:loss_prior}, and is similar to Eq.~\ref{eq:log_pdf_noise_P} in the main paper.

\section{Network Conﬁgurations}
\label{sec:supplement-netwok-configuration}
In this section, we present more details of network modules in PD-Flow.

\subsection{Augmentation Module}
\label{sec:supplement-network-augment-module}

\begin{figure}[ht]
\centering
\includegraphics[width=0.8\linewidth]{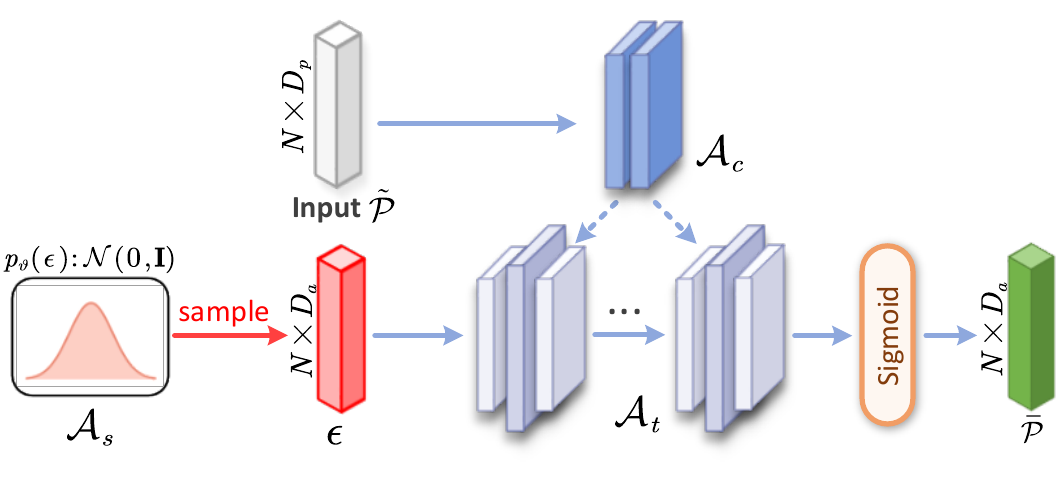}
\caption{Architecture of augmentation module $\mathcal{A}$.}
\label{fig:supplement-augment-arch}
\end{figure}

The architecture of augmentation module $\mathcal{A}$ is composed of three components, as shown in Fig.~\ref{fig:supplement-augment-arch}.

The sample module $\mathcal{A}_s$ samples random variables $\epsilon\in\mathbb{R}^{N\times D_a}$ from prior distribution $p_{\vartheta}(\epsilon)$ as initial augmented dimensions.
The condition net $\mathcal{A}_c$ is almost identical to EdgeUnit (described below), but replaces MaxPool with AvgPool.
Intuitively, $\mathcal{A}_c$ extracts neighbour context and outputs point-wise features that helps generate extra dimensions.
The transformation module $\mathcal{A}_t$ consists of three affine coupling layers and two inverse permutation layers, conditioning on features from $\mathcal{A}_c$.
Finally, a sigmoid function is applied to augmented output dimensions $\mathcal{\bar{P}}$ to avoid gradient explosion.

\subsection{Flow Components}
We briefly review the flow components used in Section \ref{sec:method_flow_module}.
The forward/inverse propagation formulas and the corresponding log-determinant $\log \left|\operatorname{det} \frac{\partial f_{\theta}^{l}}{\partial h^l}\right|$ are listed in Table~\ref{tab:flow-components}.
Note that all operations are performed on channel dimension.

{\bf Affine coupling layer.}
As the core component of flow module $\mathcal F$, the affine coupling layer, introduced in RealNVP~\cite{iclr_DinhSB17}, is a simple yet flexible paradigm to implement invertible transformation.

This layer first partitions the input $h^l$ into two parts $h_{1:d}^l$ and $h_{d:D}^l$:
$h_{1:d}^l$ maintains identity and $h_{d:D}^l$ is transformed based on $h_{1:d}^l$.
Then, we concatenate them and obtain $h^{l+1}$ as transformed output.

We show the detailed formulation in Table~\ref{tab:flow-components}, where $d$ is the partition location of channel dimension.
We set $d=\left(D_p+D_a\right) / 2$ in our experimental settings.
Transformation units $f_s^l$ and $f_b^l$ represent arbitrarily complex neural networks from $\mathbb{R}^{d} \mapsto \mathbb{R}^{D_p+D_a-d}$ and are not required to be invertible, as described in Section \ref{sec:transformation-unit}.

\begin{table}
\centering
\caption{Summarization of flow components.}
\resizebox{\linewidth}{!}{%
\begin{threeparttable}
\begin{tabular}{c|l|l|l}
\hline
Flow Component & Forward Propagation & Inverse Propagation & Log Determinant \\
\hline
\multirow{6}{*}{\makecell{Affine Coupling \\ Layer}} & $h_{1: d}^l, h_{d+1: D}^l=\textrm{split}\left(h^l\right)$ & $h_{1: d}^{l+1}, h_{d+1: D}^{l+1}=\textrm{split}\left(h^{l+1}\right)$ & \multirow{6}{*}{$\sum_{i d} h_s^l$} \\
& $h_s^l=f^l_{\theta,s} \left(h_{1: d}^l\right)$ & $h_s^l=f^l_{\theta,s} \left(h_{1: d}^{l+1}\right)$ & \\
& $h_b^l=f^l_{\theta,b}\left(h_{1: d}^l\right)$ & $h_b^l=f^l_{\theta,b}\left(h_{1: d}^{l+1}\right)$ & \\
& $h_{1: d}^{l+1}=h_{1: d}^l$ & $h_{1: d}^l=h_{1: d}^{l+1}$ & \\
& $h_{d+1: D}^{l+1}=h_{d+1: D}^l \odot \exp \left(h_s^l\right) + h_b^l$ & $h_{d+1: D}^l=\left(h_{d+1: D}^{l+1} - h_b^l\right) / \exp \left(h_s^l\right)$ & \\
& $h^{l+1}=\textrm{concat}\left(h_{1: d}^{l+1}, h_{d+1: D}^{l+1}\right)$ & $h^l=\textrm{concat}\left(h_{1: d}^l, h_{d+1: D}^l\right)$ & \\
\hline
\multirow{2}{*}{Actnorm} & \multirow{2}{*}{$h^{l+1}=\left(h^{l}-\mu\right) / \exp\left(\sigma\right)$} & \multirow{2}{*}{$h^l=h^{l + 1} \odot \exp\left(\sigma\right) +\mu$} & \multirow{2}{*}{$N\cdot\sum_{d}{\mathbf\sigma}_d$} \\  
& & & \\
\hline
\multirow{2}{*}{\makecell{Invertible $1\times 1$ \\ Convolution}} & \multirow{2}{*}{$h^{l+1}=Wh^{l}$} & \multirow{2}{*}{$h^{l}=W^{-1}h^{l + 1}$} & \multirow{2}{*}{$N \cdot \log |\det(\mathbf{W})|$} \\
& & & \\
\hline
\end{tabular}
\begin{tablenotes}
\item[*] In Log Determinant column, $i$ and $d$ denote the indices of point and channel, respectively.
\item[*] Both $\textrm{split}\left(\cdot\right)$ and $\textrm{concat}\left(\cdot\right)$ functions operate on the channel dimension.
\item[*] $\odot$ is the Hadamard product.
\end{tablenotes}
\end{threeparttable}
}
\label{tab:flow-components}
\end{table}%

{\bf Actnorm.}
The actnorm layer~\cite{kingma2018glow} applies an affine transformation to $h_i$, with trainable parameters $\mu$ and $\sigma$.
Similar to batch normalization, actnorm helps improve the training stability and performance.

In Table~\ref{tab:flow-components}, the channel-wise scale term $\mu\in\mathbb{R}^{D_p+D_a}$ and bias term $\sigma\in\mathbb{R}^{D_p+D_a}$ are initialized by the first mini-batch of data to make each channel of $h_i$ obtain zero mean and unit variance.

{\bf Permutation layer.}
Each affine coupling layer only transforms partial channels of $h_i$.
Thus, it has limited non-linear transform capability.
To ensure that all dimensions are sufficiently processed, channel permutation techniques help integrate various dimensions and improve transform diversity.

For instance, reverse and random permutations~\cite{dinh2014nice,iclr_DinhSB17} both shuffle the order of channel dimension.
The invertible $1\times 1$ convolution~\cite{kingma2018glow} (inv1x1) is a special convolutional layer that supports invertibility.
Since the dimension of $W$ is low and $\log |\det(W)|$ is relatively easy to compute, we do not apply LU decomposition~\cite{kingma2018glow} to $W$ but simply initialize $W$ randomly.
In this paper, we interchangeably use inv1x1 and inverse permutation layer.

\subsection{Transformation Unit}
\label{sec:transformation-unit}
We implement the transformation units $f_s^l$ and $f_b^l$ in the affine coupling layer in two types: {\bf EdgeUnit} and {\bf LinearUnit}.

Concretely, let $h^l\in\mathbb{R}^{N\times(D_p+D_a)}$ be the input of $l$-th affine coupling layer.
Then, we split the first $d$ channel of $h^l$ \ie $h_{1:d}^l\in\mathbb{R}^{N\times(D_p+D_a) / 2}$ as input of transformation unit $f_s^l$/$f_b^l$.

\begin{table}[h]
\centering
\caption{Architecture details of EdgeUnit.}
\setlength{\tabcolsep}{6pt}
\begin{tabular}{c|c|c|c}
\hline
 & Layer & Kernel Size & Output Size \\
\hline
K1 & Build $k$NN Graph & - & $N\times K\times 3\left(D_p+D_a\right)/2$ \\
F1 & Full-connected + ReLU & $3\left(D_p+D_a\right)/2\times D_h$ & $N\times K\times D_h$ \\
F2 & Full-connected + ReLU & $D_h\times D_h$ & $N\times K\times D_h$ \\
M1 & MaxPooling & - & $N\times D_h$ \\
F3 & Full-connected & $D_h\times\left(D_p+D_a\right)/2$ & $N\times\left(D_p+D_a\right)/2$ \\
\hline
\end{tabular}
\label{tab:arch-edgeunit}
\end{table}%

The EdgeUnit applies \emph{EdgeConv}~\cite{wang2019dynamic} to input features, which extracts 
high-level features from point-wise $k$NN graph.
The $k$NN graph extraction process is formulated as
\begin{equation}
h^{\star}_i=[h_i,\mathcal{N}\left(h_i\right),h_i-\mathcal{N}\left(h_i\right)],
\end{equation}
where $h_i=h_{1:d}^l$ are input features, $\mathcal{N}\left(\cdot\right)$ denotes the $k$NN operator,
and $h^{\star}_i\in\mathbb{R}^{K \times 3(D_p + D_a) / 2}$ indicates the point-wise features extracted from $k$NN graph.
$h^{\star}_i$ is fed into EdegUnit and then passed through the remaining layers, as listed in Table~\ref{tab:arch-edgeunit}.
The LinearUnit is simply implemented by a bunch of convolutions, as listed in Table~\ref{tab:arch-linear}.

\begin{table}[h]
\centering
\caption{Architecture details of LinearUnit.}
\setlength{\tabcolsep}{6pt}
\begin{tabular}{c|c|c|c}
\hline
 & Layer & Kernel Size & Output Size \\
\hline
C1 & Conv1D + ReLU & $\left(D_p+D_a\right)/2\times D_h$ & $N\times D_h$ \\
C2 & Conv1D + ReLU & $D_h\times D_h$ & $N\times D_h$ \\
C3 & Conv1D + ReLU & $D_h\times D_h$ & $N\times D_h$ \\
C4 & Conv1D & $D_h\times \left(D_p+D_a\right)/2$ & $N\times \left(D_p+D_a\right)/2$ \\
\hline
\end{tabular}
\label{tab:arch-linear}
\end{table}%

The flow module $\mathcal F$ transforms input $\mathcal P$ with $L=4$ to $L=12$ flow blocks.
As demonstrated in RealNVP~\cite{iclr_DinhSB17}, low levels of blocks encode high frequencies of data (\ie concrete details), whereas the high levels encode the low frequencies (\ie abstract details or basic shape).
Based on the above idea,  we use more EdgeUnit in low level blocks.
The EdgeUnit is used to extract neighbor context, whereas LinearUnit is used to extract high-level point-wise features.
For both EdgeUnit and LinearUnit, the hidden channels are set to $D_h=64$, which is the main overhead of network sizes.

\section{Implementation details}

\subsection{Dataset configuration}
\label{sec:supplement_dataset_configuration}
{\bf Training phase.}
The point clouds for training are extracted from 40 mesh models in~\cite{yu2018pu} and then split into patches of 1024 points.
These point clouds are randomly perturbed by Gaussian noise with a standard deviation of 0.5\% to 2\% of the bounding sphere’s radius.
A pair of noisy patch and the corresponding noise-free patch are cropped on the fly during training.
We adopt common data augmentation techniques for training patches, including point perturbation, scaling, and random rotation to increase data diversity and avoid over-fitting.

{\bf Evaluation phase.}
For PUSet/DMRSet, point clouds for testing are points extracted from 20/60 mesh models, at a resolution range from 10K/20K to 50K points.
Noisy test points are synthesized by adding Gaussian noise with standard deviation from 1\% to 3\% of the bounding sphere’s radius.
We normalize the noisy input points into a unit sphere before denoising and then split them into a cluster by a patch size of 1K for independent denoising.
During patch extraction, we first select seed points from input point set as patch centers by the farthest point sampling (FPS) algorithm, and grow the patch to target size by $k$NN, as done in~\cite{Luo_2021_ICCV}.
Thereafter, noisy patches are fed into PD-Flow for filtering.
Finally, we merge the output patches and sample output points to the target resolution by the FPS algorithm as our final estimation~\cite{qi2017pointnet}.
For the patch-based method, all evaluation metrics are estimated on the whole point set after merging from denoised patches.

\subsection{Network Training}
In this section, we present the training details of our method, including training strategy, network size and hyper-parameters.

For FBM filter, we set the last $D_m$ channel of $\bar{m}$ to $0$.
For LBM filter, we randomly initialize $\tilde{m}$ to $[0, 1]$ for all channels.
For LCC filter, we initialize $W$ as an identity matrix.

We train our model with Adam optimizer for 700K iterations.
The learning rate is initialized as $2\times 10^{-3}$ and updated by ReduceLROnPlateau scheduler with $\mathcal{L}_{\mathrm{EMD}}$ as the monitored metric.
We set both prior distribution $p_{\vartheta}(\tilde{z})$ and $p_{\vartheta}(\epsilon)$ as standard Gaussian distribution $\mathcal{N}(0, \mathbf{I})$.
We use the checkpoint of last training epoch as our final model.

Augmenting noisy input $\mathcal{\tilde{P}}$ with $D_a=32$ in $\mathcal A$ and using a flow module $\mathcal{F}$ with $L=8$ blocks can achieve good performance for most cases.
The tunning hyper-parameters $\alpha$, $\beta$ and $\gamma$ in Eq.~\ref{eq:total_loss} in the main paper are empirically set as $1e^{-6}$, $0.1$ and $10$, respectively.

\subsection{Partition denoising}
In experiments, we observe an obvious performance degradation of our model in denoising high resolution point clouds along with high noise levels.
This is primarily because patch-based denoise methods~\cite{pistilli2020learning,luo2020differentiable,Luo_2021_ICCV} generally use a fixed patch size as the denoising unit (\eg, 1024 points per patch).
Denoising a higher resolution point cloud indicates that the network handles a patch with a smaller surface region.
The surface estimation becomes unstable as the respective field that the network perceives become small, particularly in the denoising problem.

To resolve this problem, we first partition the high-resolution point cloud (\eg, 50K points) into several parts (\eg, 10K points), with each part sharing approximately the same shape as the original point cloud.
Then, we send each part to the network for patch-based denoising (Section \ref{sec:supplement_dataset_configuration}).
Finally, we concatenate each part back to the original resolution.
In this way, we avoid clustering much noise in a single patch and maintain good denoising performance across different point resolutions.
We only use this setting for point clouds with both high resolution and high noise levels.

Another solution to this problem may use ball-query algorithm instead of KNN to generate point patch.
This method introduces another hyperparameter (\ie radius for query), which is required to be fine-tuned for each point resolution.
From experiments, we observe that this solution does help to preserve good results under high noise level and high point resolution.
In most cases, it achieves competitive performance as the first solution.

\section{Evaluation metrics}
\label{sec:evalution_metrics}
In this section, we introduce the metrics used in quantitative comparison.
For all the metrics, the lower the values are, the better the denoising quality is.

{\bf Chamfer distance (CD)} between the predicted point cloud $\mathcal{\hat{P}}$ and ground truth point cloud $\mathcal{P}$ is defined as
\begin{equation}
\label{eq:metric_cd}
\mathcal{L}_{\text{CD}}(\mathcal{\hat{P}}, \mathcal{P})=\frac{1}{|\mathcal{\hat{P}}|} \sum_{\hat{p}\in \mathcal{\hat{P}}} \min_{p\in \mathcal{P}}\|\hat{p}-p\|+\frac{1}{\left|\mathcal{P}\right|} \sum_{p \in \mathcal{P}} \min _{\hat{p} \in \mathcal{\hat{P}}}\|p-\hat{p}\|,
\end{equation}
where $|\mathcal{\hat{P}}|$ and $|\mathcal{P}|$ denote the number of points in $\mathcal{\hat{P}}$ and $\mathcal{P}$, $\|\cdot\|$ denotes $L_2$ norm.
This first term in Eq.~\ref{eq:metric_cd} measures the average accuracy to the ground truth surface of each predicted point, whereas the second term in Eq.~\ref{eq:metric_cd} encourages an even coverage to ground truth distribution.

{\bf Point-to-mesh (P2M)} distance is defined as
\begin{equation}
\label{eq:metric_p2m}
\mathcal{L}_{\text{P2M}}(\mathcal{\hat{P}},\mathcal{M})=\frac{1}{|\mathcal{\hat{P}}|}\sum_{\hat{p} \in \mathcal{\hat{P}}} \min_{f\in\mathcal{M}}d(\hat{p},f)+\frac{1}{|\mathcal{M}|}\sum_{f\in\mathcal{M}}\min_{\hat{p}\in\mathcal{\hat{P}}}d(\hat{p}, f),
\end{equation}
where $\mathcal{M}$ is the corresponding mesh of $\mathcal{\hat{P}}$, $|\mathcal{\hat{P}}|$ is the number of points in $\mathcal{\hat{P}}$, and $f$ is the triangular face in $\mathcal{M}$ (with a total of $|\mathcal{M}|$ faces).
The $d(p,f)$ function measures the squared distance from point $p$ to face $f$. The P2M metric estimates the average accuracy that approximates the underlying surface.

{\bf Point-to-surface (P2S)} distance is defined as
\begin{equation}
\label{eq:metric_p2s}
\mathcal{L}_{\text{P2S}}(\mathcal{\hat{P}}, \mathcal{P})=\frac{1}{|\mathcal{\hat{P}}|} \sum_{\hat{p} \in \mathcal{\hat{P}}} \min_{p\in\mathcal{P}}\min_{q\in\mathcal{S}_p}\|\hat{p}-q\|,
\end{equation}
where $|\mathcal{\hat{P}}|$ denotes the number of points in $\mathcal{\hat{P}}$ and $\|\cdot\|$ denotes $L_2$ norm.
Note that $\mathcal{S}_p$ is the surface (\ie flat plane) defined by the coordinate and normal of $p$, whereas $q$ is the closest points to $\hat{p}$ on $\mathcal{S}_p$.
The P2S metric is similar to P2M metric, but requires normal data of testing point clouds instead of mesh data.

{\bf Hausdorff distance (HD)} is defined as
\begin{equation}
\label{eq:metric_hd}
\mathcal{L}_{\text{HD}}(\mathcal{\hat{P}}, \mathcal{P})=\max(\max_{\hat{p}\in\mathcal{\hat{P}}}\min_{p\in\mathcal{P}}\|\hat{p}-p\|,\max_{p\in\mathcal{P}}\min_{\hat{p}\in\mathcal{\hat{P}}}\|p-\hat{p}\|),
\end{equation}
where $\|\cdot\|$ denotes $L_2$ norm.
$\mathcal{L}_{\text{HD}}$ measures the maximum distance of point set to the nearest point in another point set, which is sensitive to outliers.

{\bf Uniform metric (Uni)}, which is proposed in PU-GAN~\cite{li2019pugan}, is used to evaluate point distribution uniformity.

This metric first uses FPS to pick seed points and then estimates the uniformity on point subset within a ball query centered at each seed point.
Depending on the ball query radius $r_d$, it can evaluate the uniformity of different area sizes.

In general, we prefer the generated point clouds to follow a uniform distribution.
In our experiment settings, we evaluate the Uni metric with $r_d=\sqrt{p}$ where $p\in\{0.4\%,0.6\%,0.8\%,1.0\%,1.2\%\}$.
Please refer to \cite{li2019pugan} for the detailed formulation.

\section{Additional Quantitative Results}
\label{sec:supplement_additional_quantitative}

\subsection{Evaluation on DMRSet}
Table~\ref{tab:comparison_sota_dmr} shows the quantitative results evaluated on DMRSet.
For a fair comparison, we retrain the deep-learning-based models including PCNet~\cite{rakotosaona2020pointcleannet}, ScoreDenoise~\cite{Luo_2021_ICCV} and our method on the training set of DMRSet, and use the pretrain model of DMRDenoise~\cite{luo2020differentiable}.
Since the resolution of 20K points is not released by authors~\cite{luo2020differentiable}, we generate 20K points and corresponding normals from the released 50K points.

\begin{table}[h]
\centering
\caption{Comparison of denoising algorithms on DMRSet.}
\resizebox{\linewidth}{!}{%
\setlength\tabcolsep{4pt}
\begin{tabular}{c| c c | c c | c c | c c | c c | c c | c c | c c }
\hline
\#Points & \multicolumn{8}{c|}{20K} & \multicolumn{8}{c}{50K} \\
\hline
Noise & \multicolumn{2}{c|}{1\%} & \multicolumn{2}{c|}{2\%} & \multicolumn{2}{c|}{2.5\%} & \multicolumn{2}{c|}{3\%} & \multicolumn{2}{c|}{1\%} & \multicolumn{2}{c|}{2\%} & \multicolumn{2}{c|}{2.5\%} & \multicolumn{2}{c}{3\%} \\
\hline
Method & \thead{CD \\ $10^{-4}$} & \thead{P2S \\ $10^{-4}$} & \thead{CD \\ $10^{-4}$} & \thead{P2S \\ $10^{-4}$} & \thead{CD \\ $10^{-4}$} & \thead{P2S \\ $10^{-4}$} & \thead{CD \\ $10^{-4}$} & \thead{P2S \\ $10^{-4}$} & \thead{CD \\ $10^{-4}$} & \thead{P2S \\ $10^{-4}$} & \thead{CD \\ $10^{-4}$} & \thead{P2S \\ $10^{-4}$} & \thead{CD \\ $10^{-4}$} & \thead{P2S \\ $10^{-4}$} & \thead{CD \\ $10^{-4}$} & \thead{P2S \\ $10^{-4}$} \\
\hline
Jet\cite{cazals2005estimating} & 2.21 & 0.56 & 2.50 & 0.73 & 2.66 & 0.84 & 2.85 & 0.98 & 2.31 & 1.01 & 4.05 & 2.39 & 4.89 & 3.08 & 5.74 & 3.78 \\
MRPCA\cite{mattei2017point} & {\bf 2.07} & {\bf 0.44} & 2.26 & {\bf 0.51} & 2.43 & {\bf 0.61} & 2.67 & {\bf 0.76} & 2.07 & 0.75 & 4.06 & 2.18 & 5.04 & 2.97 & 5.95 & 3.71 \\
GLR\cite{zeng20193d} & 2.14 & 0.53 & {\bf 2.18} & 0.59 & 2.39 & 0.73 & 2.61 & 0.89 & {\bf 1.78} & {\bf 0.65} & 2.83 & 1.47 & 3.55 & 2.04 & 4.41 & 2.72 \\
\hline
PCNet\cite{rakotosaona2020pointcleannet} & 2.25 & 0.64 & 2.69 & 0.88 & 3.02 & 1.11 & 3.36 & 1.35 & 2.15 & 0.71 & 3.43 & 1.54 & 4.18 & 2.06 & 4.92 & 2.61 \\
DMR\cite{luo2020differentiable} & 2.13 & {\bf 0.51} & 2.30 & {\bf 0.64} & 2.42 & {\bf 0.72} & {\bf 2.54} & {\bf 0.82} & {\bf 1.77} & {\bf 0.65} & 2.70 & 1.40 & 3.41 & 1.97 & 4.27 & 2.68 \\
Score\cite{Luo_2021_ICCV} & 2.21 & 0.59 & 2.48 & 0.76 & 2.61 & 0.86 & 2.74 & 0.96 & 1.91 & 0.73 & 2.81 & 1.40 & 3.43 & 1.84 & 4.19 & 2.38 \\
\hline
Ours & {\bf 2.02} & {\bf 0.50} & {\bf 2.17} & {\bf 0.61} & {\bf 2.30} & {\bf 0.71} & {\bf 2.49} & 0.86 & {\bf 1.71} & {\bf 0.65} & {\bf 2.49} & {\bf 1.15} & {\bf 2.77} & {\bf 1.45} & {\bf 3.30} & {\bf 1.92} \\
\hline
\end{tabular}
}
\label{tab:comparison_sota_dmr}
\end{table}%

From Table~\ref{tab:comparison_sota_dmr}, we can observe that our method yields better performance than other methods in most noise settings.
The performance improvement becomes obvious especially in denoising 50K points.


\subsection{Generalizability on unseen noise pattern}
In this section, we investigate the denoising performance of our method on a variety of unseen noise types.
We use the same noise settings as that of ScoreDenoise~\cite{Luo_2021_ICCV}.
Please refer to supplementary material of \cite{Luo_2021_ICCV} for detailed noise configurations.

For a fair comparison, we evaluate on the same models used in Section \ref{sec:experiment-comparison-sota}, which are trained with Gaussian noise.
We compare our method against Jet~\cite{cazals2005estimating}, MRPCA~\cite{mattei2017point}, DMRDenoise~\cite{luo2020differentiable} and ScoreDenoise~\cite{Luo_2021_ICCV}, as these methods represent the state-of-the-art performance.

Table~\ref{tab:supplement_comparison_simulate_lidar} shows the quantitative denoising results under simulated LiDAR noise.
The LiDAR noise reproduces the common noise pattern generated by scanners.
We can observe that both ScoreDenoise~\cite{Luo_2021_ICCV} and our method obtain the most accurate estimation.
Our method outperforms ScoreDenoise~\cite{Luo_2021_ICCV} in different metrics, probably thanks to the uniform distributed pattern.

\begin{table}[ht]
\centering
\caption{Comparison of denoising algorithms under simulated LiDAR noise.}
\setlength{\tabcolsep}{6pt}
\begin{tabular}{ c | c c c | c c | c }
\hline
& Jet & MRPCA & GLR & DMR & Score & Ours \\
\hline
CD  & 3.84 & 3.76 & 3.28 & 4.28 & 3.17 & {\bf 2.82} \\ \hline
P2M & 1.37 & 1.49 & 1.13 & 1.67 & 0.92 & {\bf 0.76} \\ \hline
HD  & 3.74 & 3.69 & 3.75 & 4.32 & 3.56 & {\bf 3.20} \\ \hline
\end{tabular}
\label{tab:supplement_comparison_simulate_lidar}
\end{table}%

Table~\ref{tab:supplement-comparison-noise-type} shows the quantitative denoising results under non-isotropic Gaussian noise, uni-directional noise, uniform noise, discrete noise, respectively.
Traditional methods generally rely on parameters turning to achieve good denoising performance (such as noise levels, repeating iterations and neighbor sizes, etc).
Therefore, these methods are difficult to preserve consistent performance under various noise settings and point resolutions.
Our method obtains the best results in the majority of cases.
More importantly, our method can maintain reliable quality and achieve competitive results in all metrics.
This indicates the robust generalizability of our method.

\subsection{Training on various noise distributions}

In this section, we investigate how the noise distribution used for training affects the denoising results of deep-learning based methods.
Thus, we retrain these methods on various noise types and evaluate them on PUSet with 10K points.
From Table~\ref{tab:supplement-comparison-training-set}, we observe that using the model trained on the same noise type as evaluation can further improve denoising performance.

\begin{table}
\centering
\caption{Comparison of deep-learning based methods trained on various noise types.}
\begin{tabular}{c|c| c c c | c c c | c c c }
\hline
& \makecell{Noise Type\\(Training)} & \multicolumn{3}{c|}{\makecell{isotropic\\Gaussian}} & \multicolumn{3}{c|}{\makecell{non-isotropic\\Gaussian}} & \multicolumn{3}{c}{Uniform} \\
\hline
\makecell{Noise Type\\(Evaluation)} & Method & \thead{CD \\ \scalebox{0.9}[0.9]{$10^{-4}$}} & \thead{P2M \\ \scalebox{0.9}[0.9]{$10^{-4}$}} & \thead{HD \\ \scalebox{0.9}[0.9]{$10^{-3}$}} & \thead{CD \\ \scalebox{0.9}[0.9]{$10^{-4}$}} & \thead{P2M \\ \scalebox{0.9}[0.9]{$10^{-4}$}} & \thead{HD \\ \scalebox{0.9}[0.9]{$10^{-3}$}} & \thead{CD \\ \scalebox{0.9}[0.9]{$10^{-4}$}} & \thead{P2M \\ \scalebox{0.9}[0.9]{$10^{-4}$}} & \thead{HD \\ \scalebox{0.9}[0.9]{$10^{-3}$}} \\
\hline
\multirow{3}{*}{\makecell{isotropic\\Gaussian(2\%)}} & DMR~\cite{luo2020differentiable} & 5.04 & 2.13 & 7.02 & 5.24 & 2.27 & 4.58 & 6.14 & 2.88 & 5.31 \\
& Score~\cite{Luo_2021_ICCV} & 3.68 & 1.08 & 5.78 & 3.85 & 1.34 & 6.19 & 4.86 & 1.91 & 4.56 \\
& Ours & 3.25 & 1.02 & 3.71 & 3.28 & 1.07 & 2.70 & 6.05 & 3.08 & 4.62 \\
\hline
\multirow{3}{*}{\makecell{non-isotropic\\Gaussian(2\%)}} & DMR~\cite{luo2020differentiable} & 5.26 & 2.33 & 7.40 & 4.00 & 1.31 & 3.78 & 6.24 & 2.98 & 5.40 \\
& Score~\cite{Luo_2021_ICCV} & 3.75 & 1.15 & 4.35 & 3.58 & 1.08 & 3.26 & 5.16 & 2.19 & 5.65 \\
& Ours & 3.36 & 1.12 & 3.09 & 3.19 & 1.01 & 2.87 & 6.25 & 3.29 & 6.48 \\
\hline
\multirow{3}{*}{\makecell{Uniform}} & DMR~\cite{luo2020differentiable} & 4.52 & 1.75 & 6.14 & 3.96 & 1.33 & 6.10 & 3.75 & 0.94 & 3.56 \\
& Score~\cite{Luo_2021_ICCV} & 2.48 & 0.42 & 1.27 & 3.00 & 0.92 & 1.89 & 2.43 & 0.38 & 0.92 \\
& Ours & 2.05 & 0.25 & 0.93 & 2.51 & 0.71 & 1.26 & 1.94 & 0.27 & 0.77 \\
\hline
\end{tabular}
\label{tab:supplement-comparison-training-set}
\end{table}%

\section{Additional Qualitative Results}
\label{sec:supplement_additional_qualitative}

\subsection{Visualization on denoised patch}

\begin{figure}[ht]
\includegraphics[width=\textwidth]{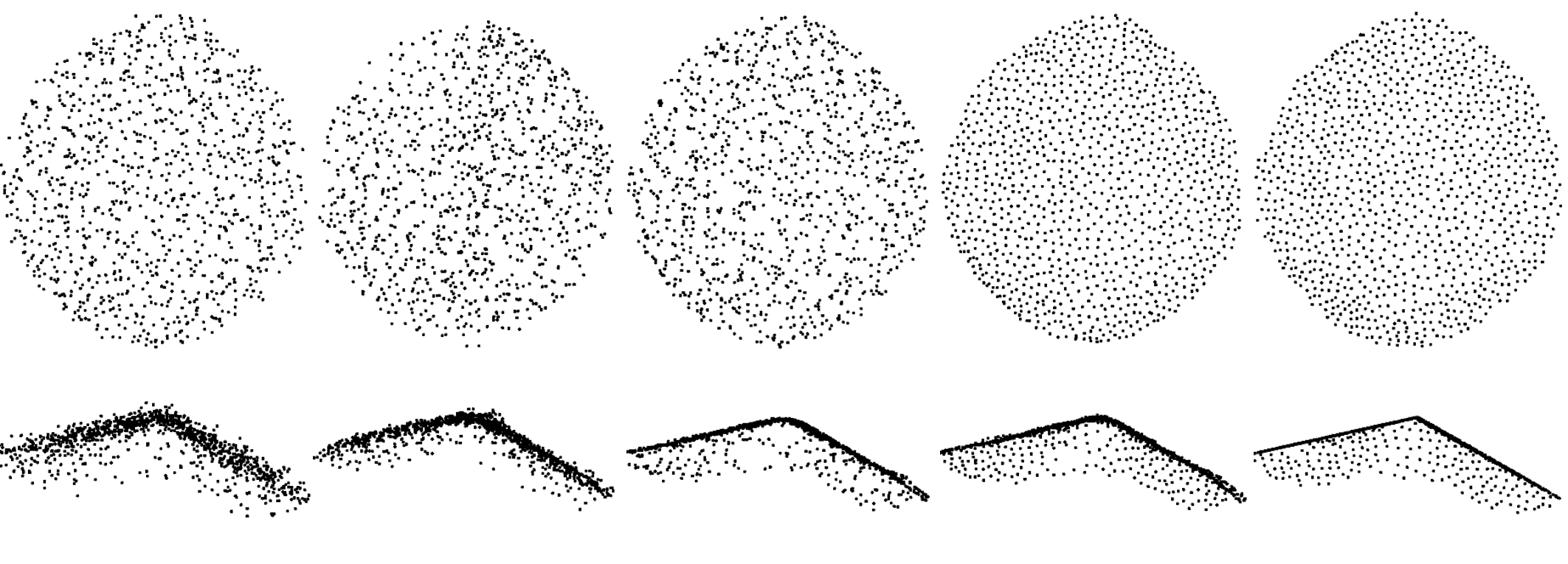}
\begin{tabularx}{\textwidth}{@{}Y@{}Y@{}Y@{}Y@{}Y@{}Y@{}Y@{}}
(a) Noisy &
(b) DMR~\cite{luo2020differentiable} &
(c) Score~\cite{Luo_2021_ICCV} &
(d) Ours &
(e) Clean
\end{tabularx}
\caption{
Visual comparison of a denoised patch from different view angles.
The first row shows the top views, and the second row shows the front views.
}
\label{fig:supple-visual-qualitative-patch}
\end{figure}

To obtain a more intuitive perception of denoised results for patch-based methods, we visualize the denoised patches of different methods from the same noisy patch (a), as shown in Fig.~\ref{fig:supple-visual-qualitative-patch}.

From the first row of Fig.~\ref{fig:supple-visual-qualitative-patch}, we observe that our method distinguishes from other methods in terms of uniformity, even though we do not explicitly enforce the uniform metric in the training loss.
From the second row of Fig.~\ref{fig:supple-visual-qualitative-patch}, both ScoreDenoise~\cite{Luo_2021_ICCV} and our method can produce points with less jitters on surface, indicating that both methods can infer smooth surface properties between estimated points.

\subsection{Investigation on disentangled latent space}

To verify whether the latent point representation is disentangled, we conduct an experiment to investigate how the latent $z$ of NFs affects generative quality.
We inspect the effect of clean and noisy channels by adding noise $o_d$.

\begin{figure}[ht]
\centering
\includegraphics[width=\linewidth]{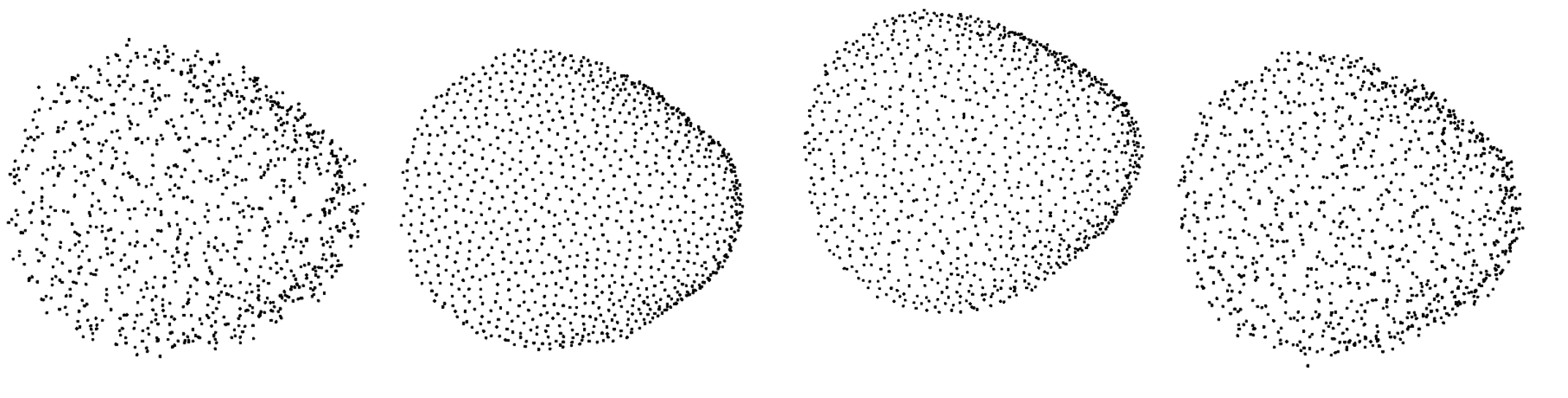}
\begin{tabularx}{\linewidth}{@{}Y@{}Y@{}X@{}X@{}}
(a) $\tilde{z}=[z_p,z_n]$ & (b) $\hat{z}=[z_p, z_c]$ & (c) $z_1=[z_p + o_d, z_c]$ & (d) $z_2=[z_p, z_c + o_d]$ \\
\end{tabularx}
\caption{
Point patches generated from different latent vectors $\hat{z}$.
}
\label{fig:supple-visual-patch-experiment}
\end{figure}

To be specific, let $z_p$ and $z_n$ be the channels of noisy latent $\tilde{z}$, which are supposed to embed point and noise respectively.
We set $D_a=33$ and use FBM filter with $\bar{m} = [\underbrace{1,1,\cdots,1}_\textrm{18 elements}, \underbrace{0,0,\cdots,0}_\textrm{18 elements}]$, such that $z_p$ and $z_n$ both contain 18 channels.
Let $z_c\in\mathbb{R}^{18}$ denotes the channels denoised by FBM filter, where $z_c=0$ in this case.
Fig.~\ref{fig:supple-visual-patch-experiment} visualizes the patches decoded from several latent $z$ combinations by adding noise to different components.
The directional noise $o_d$ is defined as $o_d=\delta\mathbf{1}$, where $\mathbf{1}\in\mathbb{R}^{18}$ is an all-ones vector and $\delta$ is a scalar adjusted by user ($\delta=1.0$ in this case).

Since Fig.~\ref{fig:supple-visual-patch-experiment}a does not change the noisy latent $\tilde{z}$, our method generates the same patch as noisy input due to the invertibility of NF.
By replacing $z_n$ with $z_c$, we obtain a well-distributed patch in Fig.~\ref{fig:supple-visual-patch-experiment}b.
Since $z_c$ does not embed any information and NF is a lossless propagation process, without loss of generality, we can hypothesize that NF encodes the noisy patch to a regularized space of $z_p$ where points are uniformly distributed and embeds the point-wise distortion to $z_n$.
This gives us an intuitive explanation of why our method achieves better uniformity.

If we add directional noise $o_d$ to $z_p$, we observe an unified translation to each point in Fig.~\ref{fig:supple-visual-patch-experiment}c.
For figure plotting convenience, we only show a relative small amount of translation in Fig.~\ref{fig:supple-visual-patch-experiment}c.
This phenomenon indicates that $z_p$ probably embeds some information that is absolute to point's coordinate.
If we add directional noise $o_d$ to $z_c$, we observe that each point in Fig.~\ref{fig:supple-visual-patch-experiment}d is corrupted by random noise, indicating that the $z_c$ channels probably encode some properties that disturb patch uniformity.
The different behaviors in Fig.~\ref{fig:supple-visual-patch-experiment}c and Fig.~\ref{fig:supple-visual-patch-experiment}d verify that $z_p$ and $z_c$ share different functions in latent point representation; in other words, the latent code space of $z$ is disentangled.


\subsection{More denoised results}

\begin{figure}
\centering
\ifdefined\GENERALCOMPILE
    \ifdefined\HIGHRESOLUTIONFIGURE
        \includegraphics[width=\textwidth]{images/plotting/RueMadame/RueMadame46-high-resolution.pdf}
    \else
        \includegraphics[width=\textwidth]{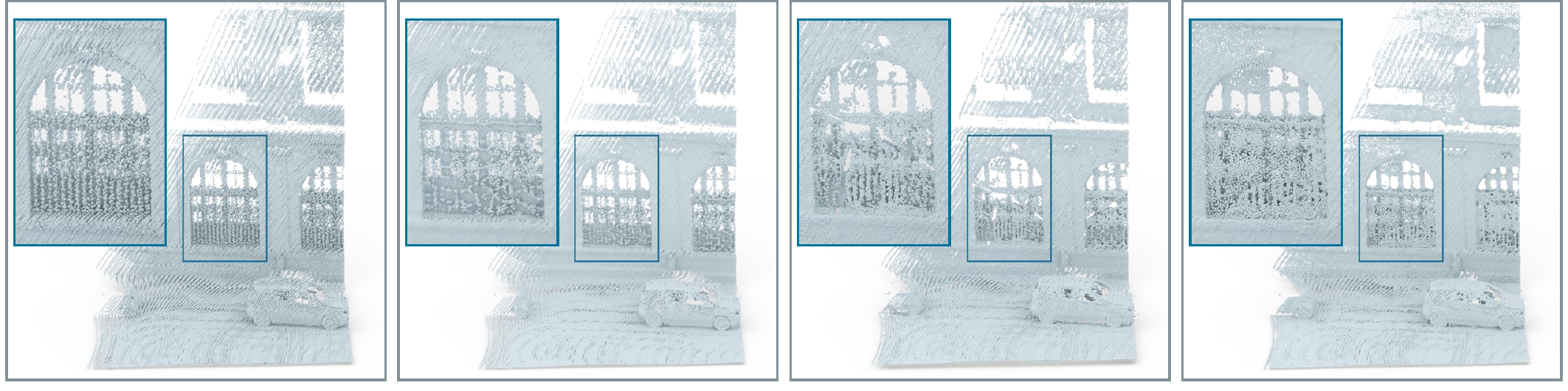}
    \fi
\fi

\ifdefined\GENERALCOMPILE
    \ifdefined\HIGHRESOLUTIONFIGURE
        \includegraphics[width=\textwidth]{images/plotting/RueMadame/RueMadame56-high-resolution.pdf}
    \else
        \includegraphics[width=\textwidth]{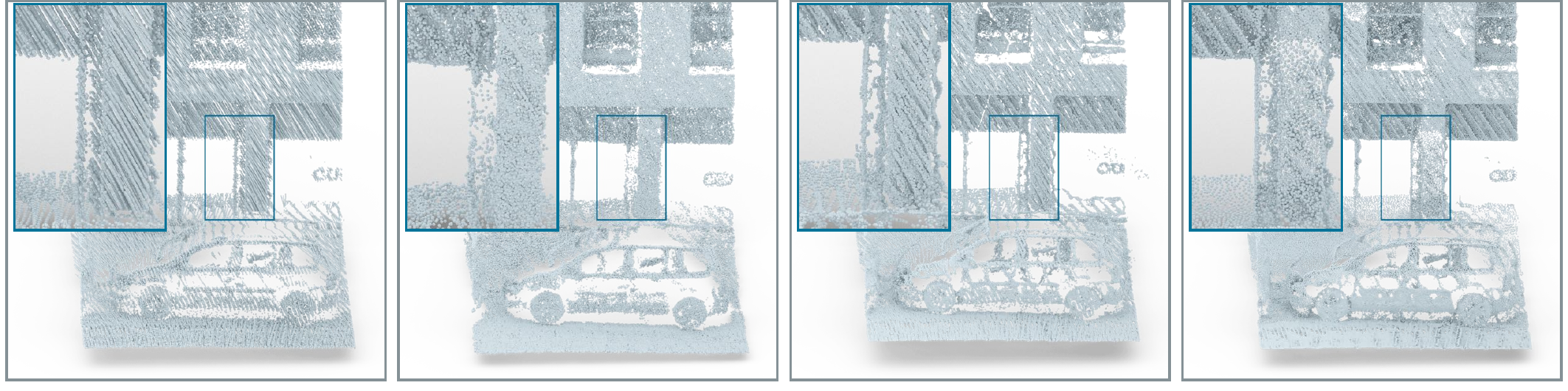}
    \fi
\fi

\begin{tabularx}{\textwidth}{@{}Y@{}Y@{}Y@{}Y@{}@{}}
(a) Noisy &
(b) DMRDenoise~\cite{luo2020differentiable} &
(c) ScoreDenoise~\cite{Luo_2021_ICCV} &
(d) Ours
\end{tabularx}

\caption{More visual comparison on \emph{Paris-rue-Madame}~\cite{serna2014paris} dataset.}
\label{fig:supple-visual-more-ruemadame}
\end{figure}

\begin{figure}[!p]
\centering
\includegraphics[width=0.85\textwidth]{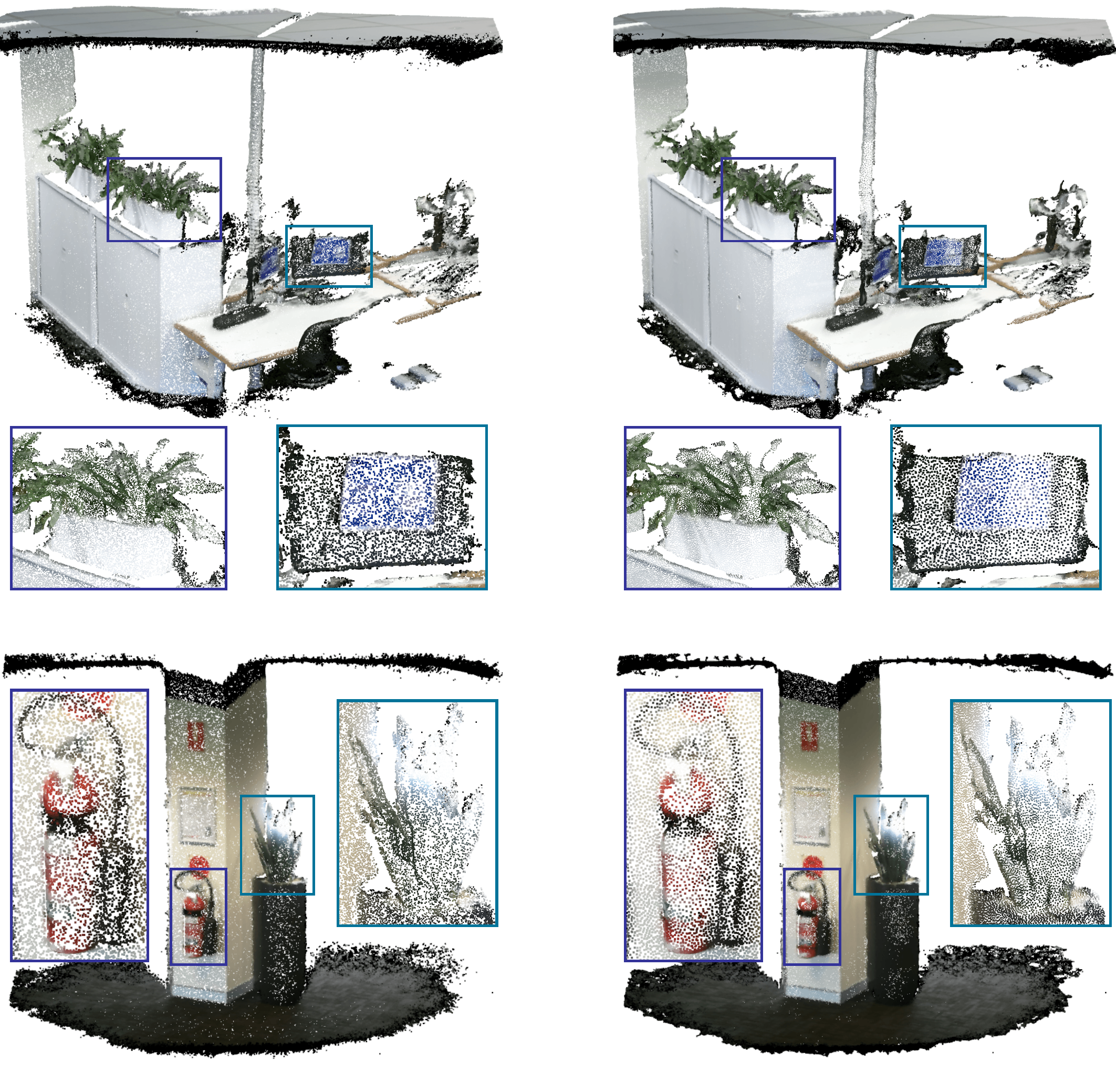}
\vspace{-10pt}
\begin{tabularx}{\textwidth}{@{}Y@{}Y@{}}
(a) Noisy & (b) Ours \\
\end{tabularx}
\caption{Denoised results on 3DCSR~\cite{huang2021comprehensive} dataset.}
\label{fig:supple-visual-3DCSR}
\end{figure}

\begin{figure}[!t]
\ifdefined\GENERALCOMPILE
    \ifdefined\HIGHRESOLUTIONFIGURE
        \includegraphics[width=\textwidth]{images/plotting/noise/LiDAR-high-resolution2.pdf}
    \else
        \includegraphics[width=\textwidth]{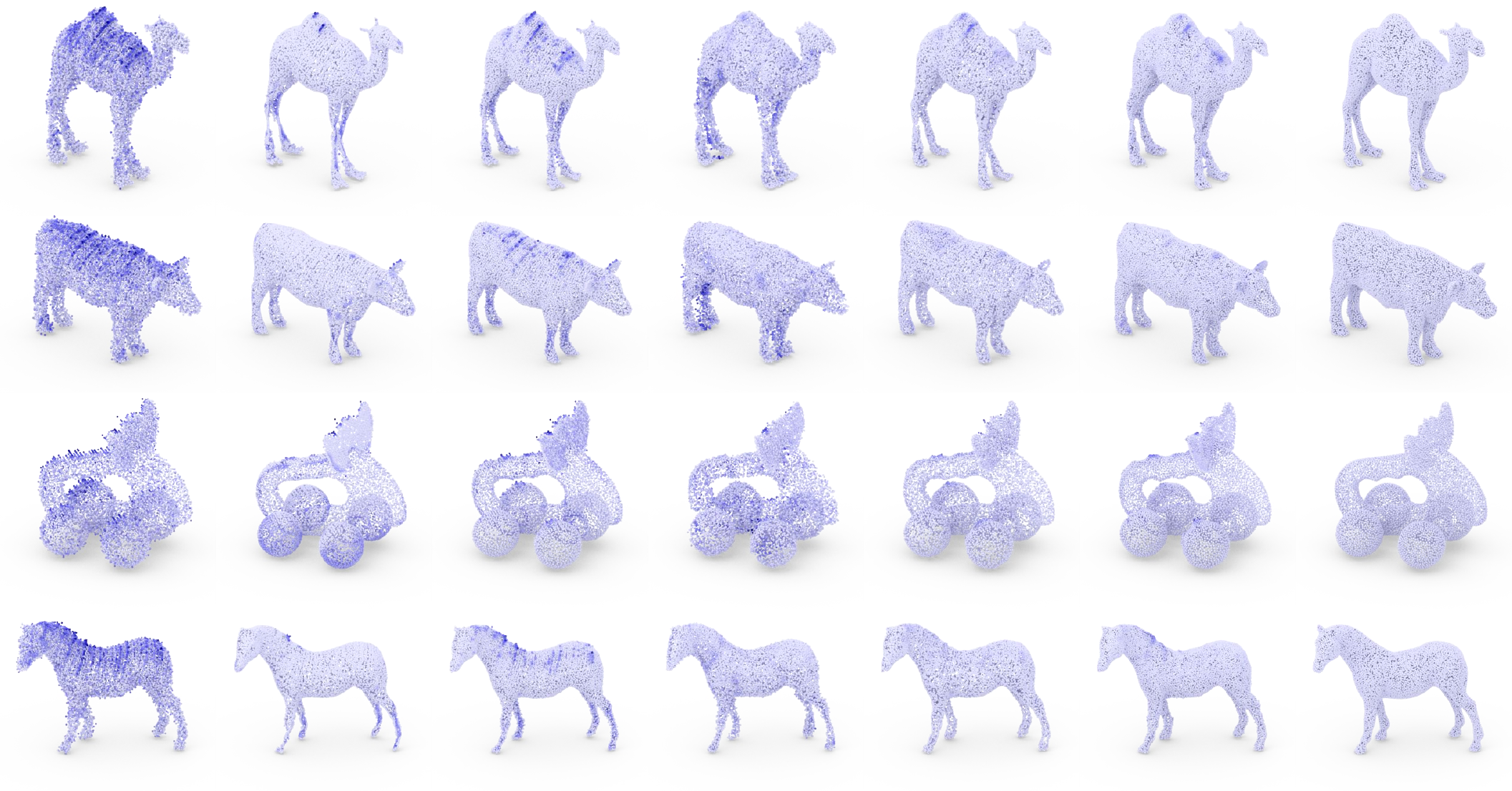}
    \fi
\fi
\begin{tabularx}{\textwidth}{@{}Y@{}Y@{}Y@{}Y@{}Y@{}Y@{}Y@{}}
\scalebox{0.8}[0.8]{(a) Noisy} &
\scalebox{0.8}[0.8]{(b)MRPCA\cite{mattei2017point}} &
\scalebox{0.8}[0.8]{(c) PCNet\cite{rakotosaona2020pointcleannet}} &
\scalebox{0.8}[0.8]{(d) DMR \cite{luo2020differentiable}} &
\scalebox{0.8}[0.8]{(e) Score \cite{Luo_2021_ICCV}} &
\scalebox{0.8}[0.8]{(f) Ours} &
\scalebox{0.8}[0.8]{(g) Clean}
\end{tabularx}
\caption{Visual comparison under simulated LiDAR noise.}
\label{fig:supple-visual-LiDAR-noise}
\end{figure}

\begin{figure}
\centering
\captionsetup[subfloat]{farskip=2pt,captionskip=1pt,position=top}
\par\noindent\rule{0.95\textwidth}{0.4pt}
\subfloat[Gaussian 1\%]{%
\ifdefined\HIGHRESOLUTIONFIGURE
    \includegraphics[width=0.95\linewidth, align=c]{images/plotting/noise/noise001-high-resolution.pdf}%
\else
    \includegraphics[width=0.95\linewidth, align=c]{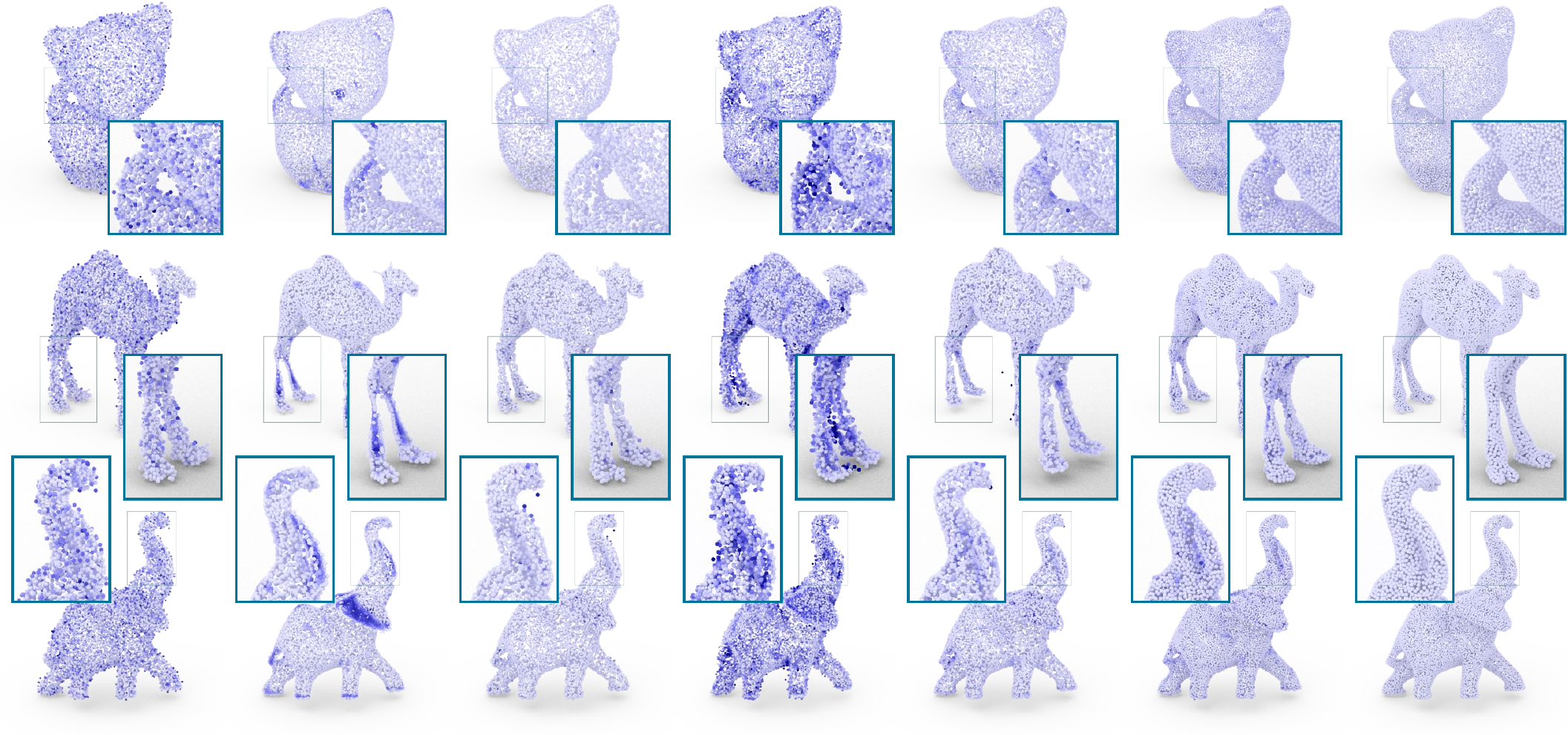}%
\fi
}%
\\
\par\noindent\rule{0.95\textwidth}{0.4pt}
\subfloat[Gaussian 2\%]{%
\ifdefined\HIGHRESOLUTIONFIGURE
    \includegraphics[width=0.95\linewidth, align=c]{images/plotting/noise/noise002-high-resolution.pdf}%
\else
    \includegraphics[width=0.95\linewidth, align=c]{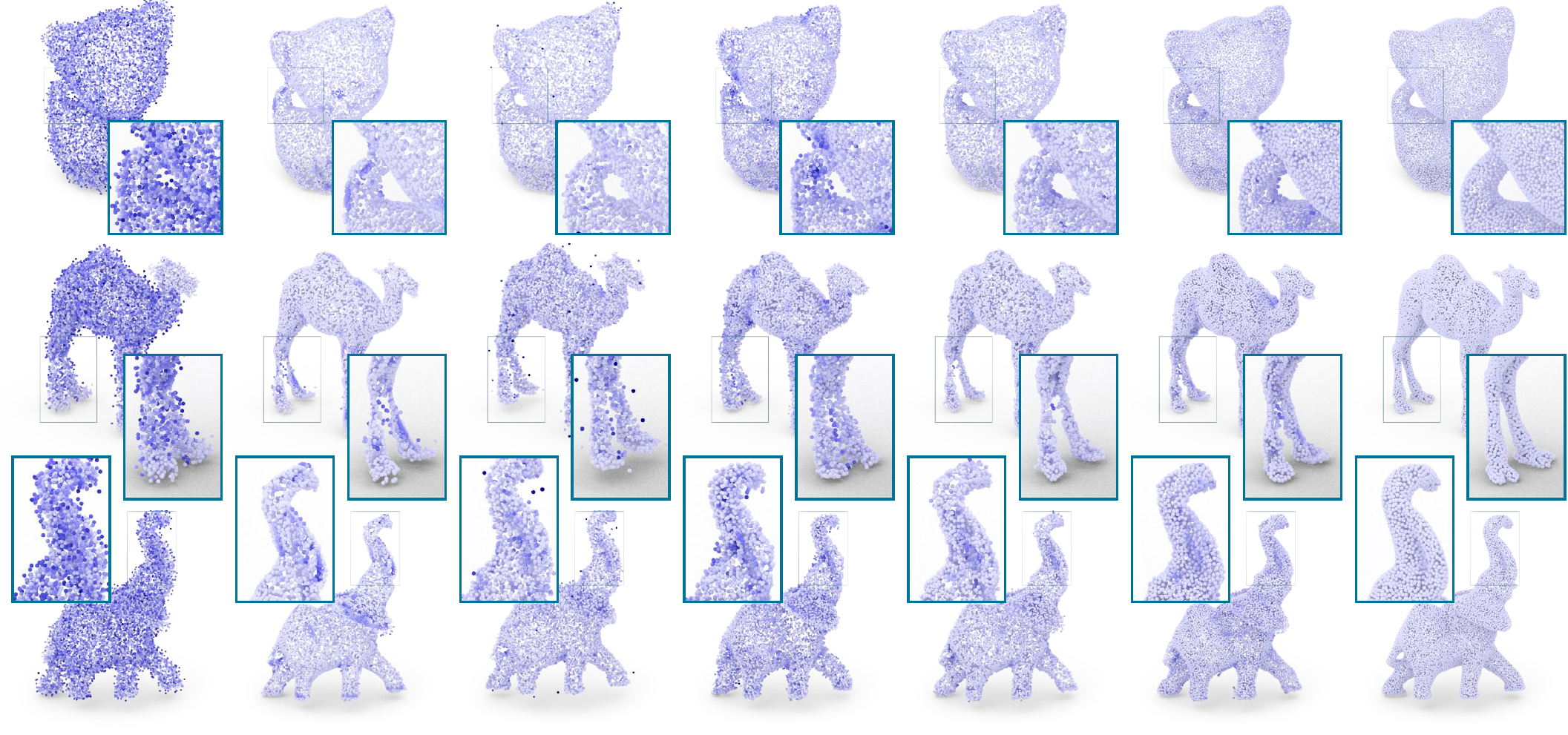}%
\fi
}%
\\
\par\noindent\rule{0.95\textwidth}{0.4pt}
\subfloat[Gaussian 3\%]{%
\ifdefined\HIGHRESOLUTIONFIGURE
    \includegraphics[width=0.95\linewidth, align=c]{images/plotting/noise/noise003-high-resolution.pdf}%
\else
    \includegraphics[width=0.95\linewidth, align=c]{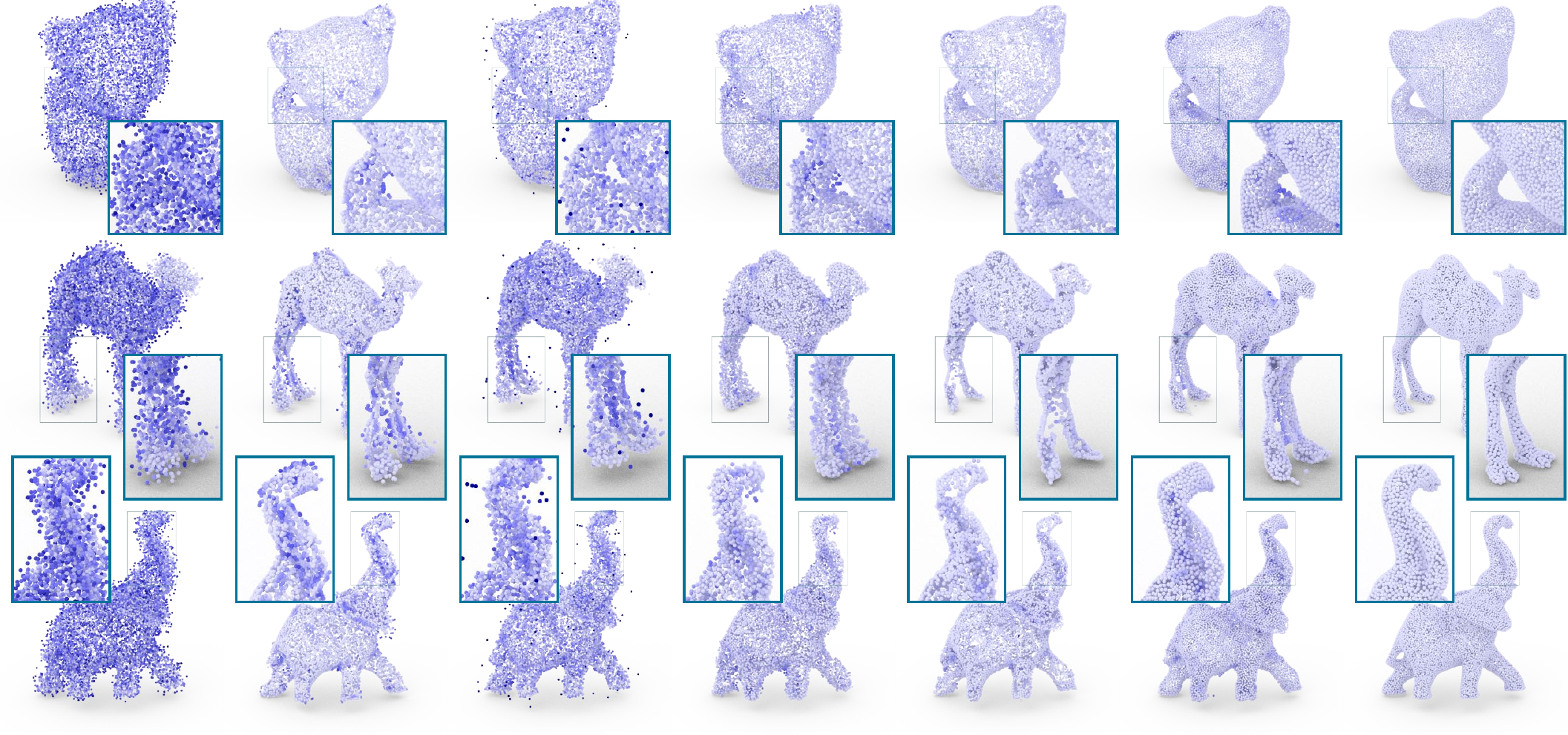}%
\fi
}%
\par\noindent\rule{0.95\textwidth}{0.4pt}

\begin{tabularx}{0.95\textwidth}{@{}Y@{}Y@{}Y@{}Y@{}Y@{}Y@{}Y@{}}
Noisy &
\scalebox{0.9}[0.9]{MRPCA~\cite{mattei2017point}} &
\scalebox{0.9}[0.9]{PCNet~\cite{rakotosaona2020pointcleannet}} &
\scalebox{0.9}[0.9]{DMR~\cite{luo2020differentiable}} &
\scalebox{0.9}[0.9]{Score~\cite{Luo_2021_ICCV}} &
\scalebox{0.9}[0.9]{Ours} &
\scalebox{0.9}[0.9]{Clean}
\end{tabularx}

\caption{
Visual comparison under various noise levels.
}
\label{fig:supple-visual-noise-levels}
\end{figure}

\begin{figure}
\centering
\captionsetup[subfloat]{farskip=5pt,captionskip=5pt,position=top}
\par\noindent\rule{0.95\textwidth}{0.4pt}
\subfloat[Non-isotropic Gaussian Noise]{%
\ifdefined\HIGHRESOLUTIONFIGURE
    \includegraphics[width=0.98\linewidth, align=c]{images/plotting/noise/non-isotropic-gaussian-high-resolution.pdf}%
\else
    \includegraphics[width=0.98\linewidth, align=c]{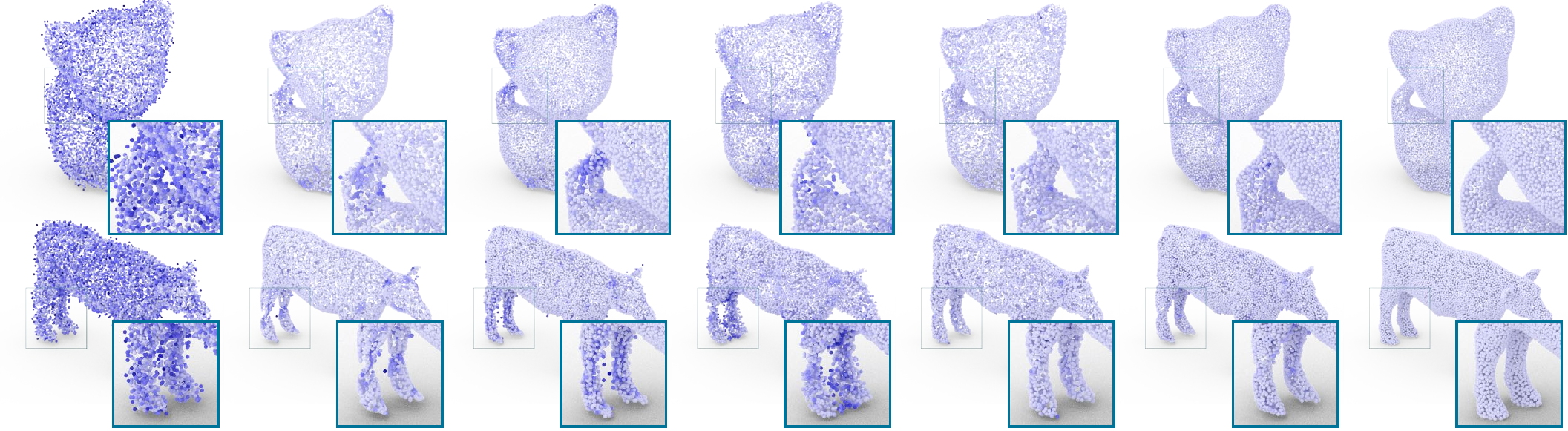}%
\fi
}%
\\
\par\noindent\rule{0.95\textwidth}{0.4pt}
\subfloat[Uni-directional Noise]{%
\ifdefined\HIGHRESOLUTIONFIGURE
    \includegraphics[width=0.98\linewidth, align=c]{images/plotting/noise/uni-directional-high-resolution.pdf}%
\else
    \includegraphics[width=0.98\linewidth, align=c]{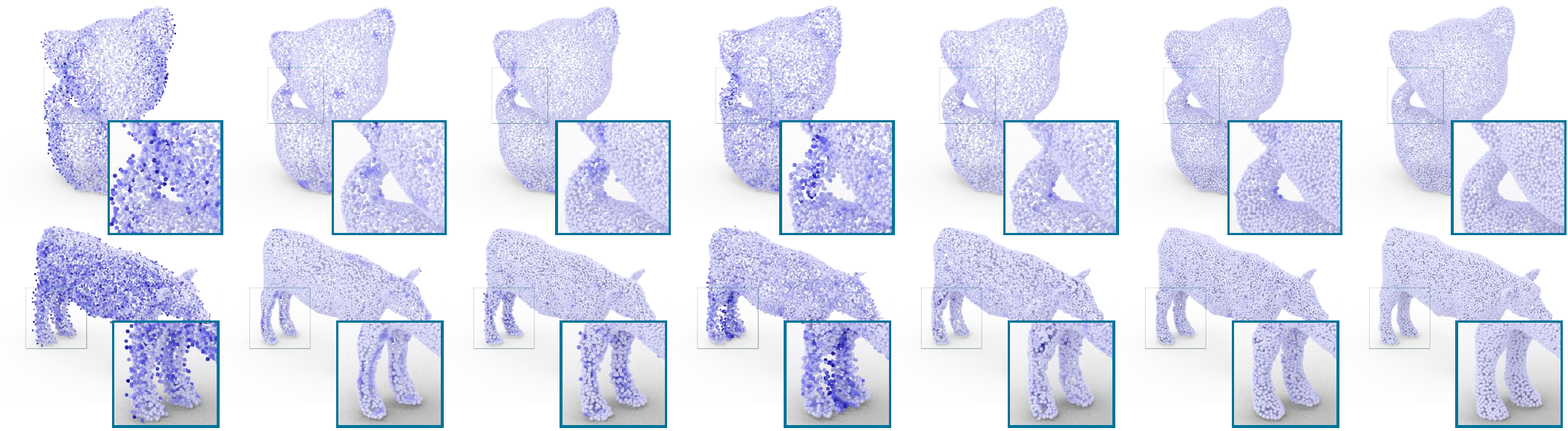}%
\fi
}%
\\
\par\noindent\rule{0.95\textwidth}{0.4pt}
\subfloat[Uniform Noise]{%
\ifdefined\HIGHRESOLUTIONFIGURE
    \includegraphics[width=0.98\linewidth, align=c]{images/plotting/noise/uniform-high-resolution.pdf}%
\else
    \includegraphics[width=0.98\linewidth, align=c]{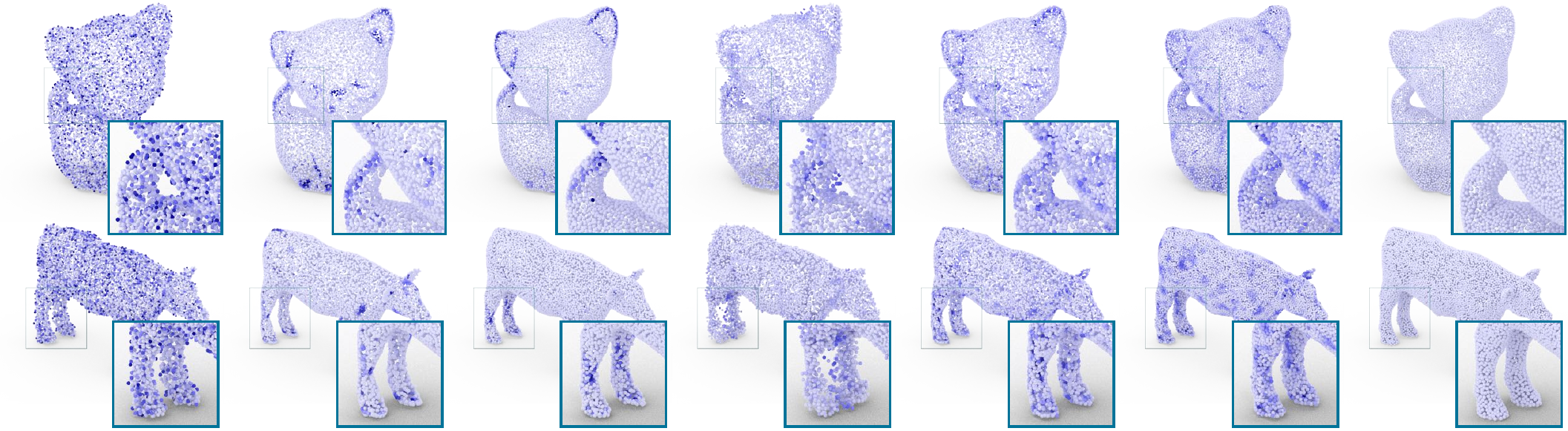}%
\fi
}%
\\
\par\noindent\rule{0.95\textwidth}{0.4pt}
\subfloat[Discrete Noise]{%
\ifdefined\HIGHRESOLUTIONFIGURE
    \includegraphics[width=0.98\linewidth, align=c]{images/plotting/noise/discrete-high-resolution.pdf}%
\else
    \includegraphics[width=0.98\linewidth, align=c]{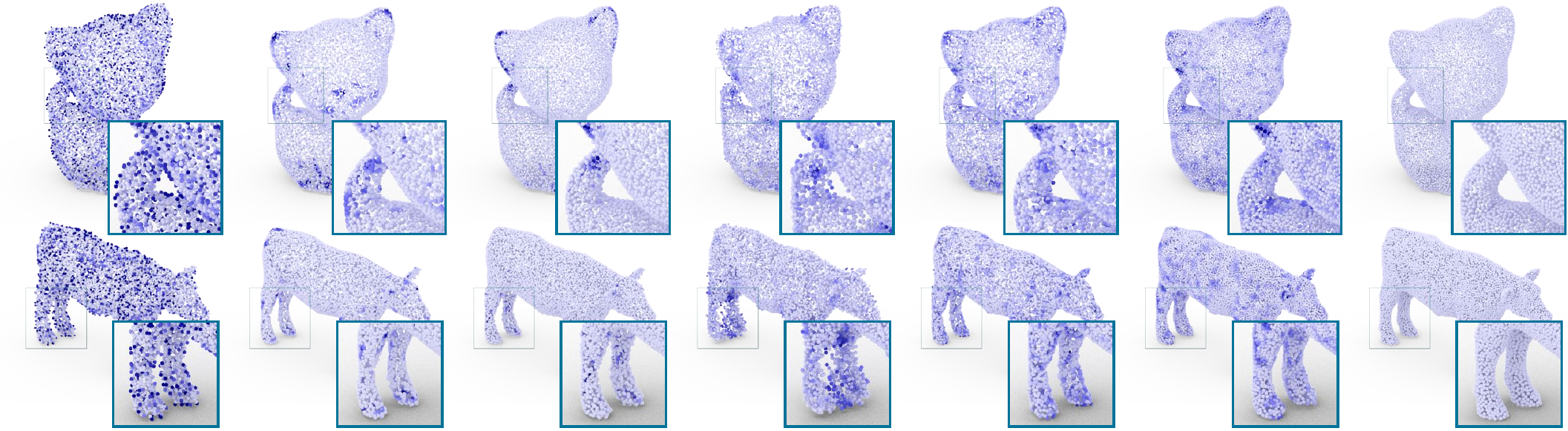}%
\fi
}%
\par\noindent\rule{0.95\textwidth}{0.4pt}

\begin{tabularx}{0.98\textwidth}{@{}Y@{}Y@{}Y@{}Y@{}Y@{}Y@{}Y@{}}
Noisy &
\scalebox{0.9}[0.9]{MRPCA~\cite{mattei2017point}} &
\scalebox{0.9}[0.9]{PCNet~\cite{rakotosaona2020pointcleannet}} &
\scalebox{0.9}[0.9]{DMR~\cite{luo2020differentiable}} &
\scalebox{0.9}[0.9]{Score~\cite{Luo_2021_ICCV}} &
\scalebox{0.9}[0.9]{Ours} &
\scalebox{0.9}[0.9]{Clean}
\end{tabularx}

\caption{Visual comparison under various noise types.}
\label{fig:supple-visual-noise-types}
\end{figure}

In this section, we illustrate more denoised results in the following figures.
Fig.~\ref{fig:supple-visual-Paris-CARLA-3D} shows some denoising examples on Paris-CARLA-3D~\cite{deschaud2021pariscarla3d} dataset.
Fig.~\ref{fig:supple-visual-more-ruemadame} shows more denoising examples on \emph{Paris-rue-Madame}~\cite{serna2014paris} dataset.
Fig.~\ref{fig:supple-visual-3DCSR} shows some denoising examples on 3DCSR~\cite{huang2021comprehensive} dataset.
Fig.~\ref{fig:supple-visual-LiDAR-noise} shows more denoising examples under simulated LiDAR noise.
Fig.~\ref{fig:supple-visual-noise-levels} shows more denoising examples under 1\% to 3\% Gaussian levels.
Fig.~\ref{fig:supple-visual-noise-types} shows more denoising examples under unseen noise types.

\section{Limitation and Future work}


Most existing methods (including our method) do not contain noise level estimation.
Therefore, how many denoising iterations are needed to achieve the best result is uncertain.
Generally, for our method, a single iteration is enough for low noise level (\eg, 1\% and 2\% Gaussian noise), while 2 iterations are needed for high noise level (\eg, 3\% Gaussian noise).


In the future, we will enhance the denoising capability of PD-Flow by introducing the outlier filtering operator and make further investigation on the influence of latent dimensions.
Furthermore, we can extend the propagation pipeline of PD-Flow to point cloud compression task, which takes high priority in efficient storage and detail reconstruction.

\newpage

\begin{sidewaystable}
\centering
\caption{Comparison of denoising algorithms under various noise types on PUSet.}
\begin{tabular}{c|c| c c c | c c c | c c c || c c c | c c c | c c c }
\hline
\multicolumn{2}{c|}{\#Points} & \multicolumn{9}{c||}{10K} & \multicolumn{9}{c}{50K} \\
\hline
\multicolumn{2}{c|}{Noise Level} & \multicolumn{3}{c|}{1\%} & \multicolumn{3}{c|}{2\%} & \multicolumn{3}{c||}{3\%} & \multicolumn{3}{c|}{1\%} & \multicolumn{3}{c|}{2\%} & \multicolumn{3}{c}{3\%} \\
\hline
Noise Type & Method & \thead{CD \\ \scalebox{0.9}[0.9]{$10^{-4}$}} & \thead{P2M \\ \scalebox{0.9}[0.9]{$10^{-4}$}} & \thead{HD \\ \scalebox{0.9}[0.9]{$10^{-3}$}} & \thead{CD \\ \scalebox{0.9}[0.9]{$10^{-4}$}} & \thead{P2M \\ \scalebox{0.9}[0.9]{$10^{-4}$}} & \thead{HD \\ \scalebox{0.9}[0.9]{$10^{-3}$}} & \thead{CD \\ \scalebox{0.9}[0.9]{$10^{-4}$}} & \thead{P2M \\ \scalebox{0.9}[0.9]{$10^{-4}$}} & \thead{HD \\ \scalebox{0.9}[0.9]{$10^{-3}$}} & \thead{CD \\ \scalebox{0.9}[0.9]{$10^{-4}$}} & \thead{P2M \\ \scalebox{0.9}[0.9]{$10^{-4}$}} & \thead{HD \\ \scalebox{0.9}[0.9]{$10^{-3}$}} & \thead{CD \\ \scalebox{0.9}[0.9]{$10^{-4}$}} & \thead{P2M \\ \scalebox{0.9}[0.9]{$10^{-4}$}} & \thead{HD \\ \scalebox{0.9}[0.9]{$10^{-3}$}} & \thead{CD \\ \scalebox{0.9}[0.9]{$10^{-4}$}} & \thead{P2M \\ \scalebox{0.9}[0.9]{$10^{-4}$}} & \thead{HD \\ \scalebox{0.9}[0.9]{$10^{-3}$}} \\
\hline
\multirow{5}{*}{\makecell{non-isotropic\\Gaussian}} & Jet~\cite{cazals2005estimating} & 3.30 & 1.07 & 1.81 & 4.71 & 1.82 & 3.21 & 6.25 & 3.01 & 9.49 & 0.93 & 0.26 & 1.05 & 1.86 & 0.94 & 4.49 & 4.02 & 2.81 & 10.6 \\
& MRPCA~\cite{mattei2017point} & 3.35 & 1.30 & 1.71 & 4.06 & 1.48 & {\bf 2.84} & {\bf 4.90} & {\bf 2.06} & 6.85 & 0.70 & {\bf 0.13} & {\bf 0.77} & 1.51 & 0.70 & 3.92 & 4.61 & 3.22 & 10.3 \\
& DMR~\cite{luo2020differentiable} & 4.63 & 1.82 & 5.46 & 5.26 & 2.33 & 7.40 & 6.29 & 3.20 & 10.1 & 1.22 & 0.50 & 2.76 & 1.70 & 0.91 & 4.26 & 2.78 & 1.85 & 6.79 \\
& Score~\cite{Luo_2021_ICCV} & 2.50 & 0.47 & 1.57 & 3.75 & 1.15 & 4.35 & {\bf 4.90} & 2.14 & 10.0 & 0.72 & 0.15 & 2.85 & 1.38 & {\bf 0.65} & 5.49 & {\bf 2.24} & {\bf 1.31} & 6.24 \\
\cline{2-20}
& Ours & {\bf 2.13} & {\bf 0.40} & {\bf 1.16} & {\bf 3.36} & {\bf 1.12} & {\bf 3.09} & 5.01 & 2.56 & {\bf 5.42} & {\bf 0.66} & 0.17 & {\bf 1.04} & {\bf 1.22} & {\bf 0.63} & {\bf 2.17} & {\bf 2.30} & 1.60 & {\bf 5.12} \\
\hline
\hline
\multirow{5}{*}{\makecell{Uni-\\directional}} & Jet~\cite{cazals2005estimating} & 2.16 & 0.88 & 1.50 & 3.29 & 1.23 & 2.31 & 4.20 & 1.75 & 5.21 & 0.64 & 0.17 & 0.65 & 1.01 & 0.37 & 2.08 & 1.69 & 0.90 & 6.38 \\
& MRPCA~\cite{mattei2017point} & 2.43 & 1.25 & 1.64 & 3.29 & 1.43 & 2.16 & 3.91 & 1.71 & 3.96 & 0.52 & 0.11 & {\bf 0.52} & {\bf 0.78} & {\bf 0.23} & 2.17 & 1.66 & 0.91 & 6.30 \\
& DMR~\cite{luo2020differentiable} & 4.47 & 1.75 & 6.06 & 4.75 & 1.93 & 5.87 & 5.25 & 2.35 & 7.04 & 1.08 & 0.40 & 2.05 & 1.26 & 0.54 & 2.79 & 1.64 & 0.86 & 4.99 \\
& Score~\cite{Luo_2021_ICCV} & 1.47 & 0.28 & 1.36 & 2.46 & 0.56 & 2.13 & 3.50 & 1.25 & 5.77 & {\bf 0.49} & {\bf 0.06} & 1.28 & {\bf 0.81} & 0.26 & 2.47 & {\bf 1.19} & {\bf 0.55} & 9.48 \\
\cline{2-20}
& Ours & {\bf 1.19} & {\bf 0.23} & {\bf 0.75} & {\bf 2.21} & {\bf 0.51} & {\bf 1.74} & {\bf 3.07} & {\bf 1.06} & {\bf 3.77} & 0.47 & {\bf 0.08} & {\bf 0.58} & {\bf 0.81} & 0.32 & {\bf 1.31} & 1.30 & 0.71 & {\bf 2.70} \\
\hline
\hline
\multirow{5}{*}{Uniform} & Jet~\cite{cazals2005estimating} & 1.97 & 0.83 & 1.23 & 3.30 & 1.02 & 1.49 & 4.04 & 1.33 & 2.01 & 0.65 & 0.13 & 0.33 & 0.89 & 0.23 & {\bf 0.56} & 1.20 & 0.42 & 0.91 \\
& MRPCA~\cite{mattei2017point} & 2.29 & 1.24 & 1.47 & 3.40 & 1.28 & 1.69 & 3.82 & 1.35 & 2.03 & 0.61 & 0.12 & 0.33 & {\bf 0.52} & {\bf 0.11} & 0.61 & 0.86 & {\bf 0.22} & {\bf 0.80} \\
& DMR~\cite{luo2020differentiable} & 4.42 & 1.70 & 6.17 & 4.52 & 1.75 & 6.14 & 4.75 & 1.93 & 6.61 & 1.09 & 0.40 & 1.65 & 1.21 & 0.50 & 2.12 & 1.33 & 0.60 & 2.87 \\
& Score~\cite{Luo_2021_ICCV} & 1.31 & 0.25 & 0.72 & 2.48 & 0.42 & 1.27 & 3.41 & 1.01 & 7.30 & 0.50 & {\bf 0.04} & 1.19 & 0.70 & 0.13 & 1.67 & 0.91 & 0.30 & 2.90 \\
\cline{2-20}
& Ours & {\bf 0.92} & {\bf 0.18} & {\bf 0.61} & {\bf 2.05} & {\bf 0.35} & {\bf 0.93} & {\bf 2.70} & {\bf 0.68} & {\bf 1.44} & {\bf 0.45} & {\bf 0.05} & {\bf 0.29} & {\bf 0.63} & 0.15 & {\bf 0.65} & {\bf 0.89} & 0.36 & {\bf 1.22} \\
\hline
\hline
\multirow{5}{*}{Discrete} & Jet~\cite{cazals2005estimating} & 1.93 & 0.82 & 1.32 & 2.89 & 0.98 & 1.52 & 3.40 & 1.26 & 1.90 & 0.56 & 0.14 & 0.34 & 0.91 & 0.15 & {\bf 0.44} & 1.08 & 0.43 & 0.99 \\
& MRPCA~\cite{mattei2017point} & 2.25 & 1.24 & 1.49 & 3.02 & 1.28 & 1.67 & 3.22 & 1.32 & 1.86 & 1.11 & 0.46 & 0.92 & {\bf 0.54} & {\bf 0.11} & 0.47 & {\bf 0.61} & {\bf 0.15} & {\bf 0.80} \\
& DMR~\cite{luo2020differentiable} & 4.42 & 1.70 & 5.85 & 4.58 & 1.78 & 5.25 & 4.84 & 1.98 & 6.61 & 1.11 & 0.42 & 1.83 & 1.21 & 0.50 & 2.36 & 1.37 & 0.63 & 2.77 \\
& Score~\cite{Luo_2021_ICCV} & 1.27 & 0.25 & 0.93 & 2.19 & 0.42 & 4.27 & 3.08 & 1.01 & 3.07 & {\bf 0.45} & {\bf 0.05} & 1.20 & 0.62 & 0.14 & 1.84 & 0.78 & 0.25 & 2.02 \\
\cline{2-20}
& Ours & {\bf 0.91} & {\bf 0.18} & {\bf 0.60} & {\bf 1.90} & {\bf 0.35} & {\bf 0.91} & {\bf 2.58} & {\bf 0.69} & {\bf 1.31} & {\bf 0.43} & {\bf 0.06} & {\bf 0.34} & {\bf 0.59} & 0.14 & {\bf 0.96} & 0.83 & 0.32 & 1.45 \\
\hline
\end{tabular}
\label{tab:supplement-comparison-noise-type}
\end{sidewaystable}%

\clearpage
\newpage

\begin{figure}[!p]
\centering
\includegraphics[width=\textwidth]{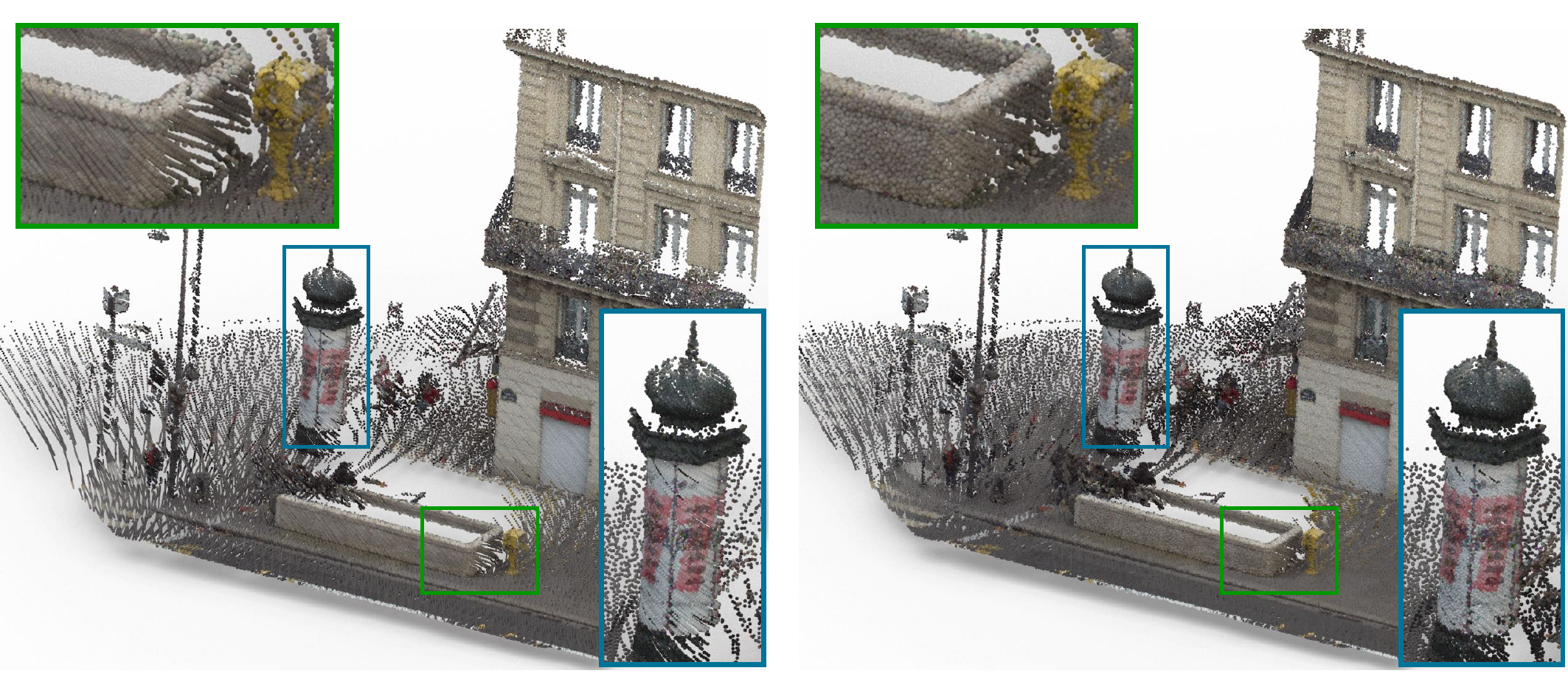}
\includegraphics[width=\textwidth]{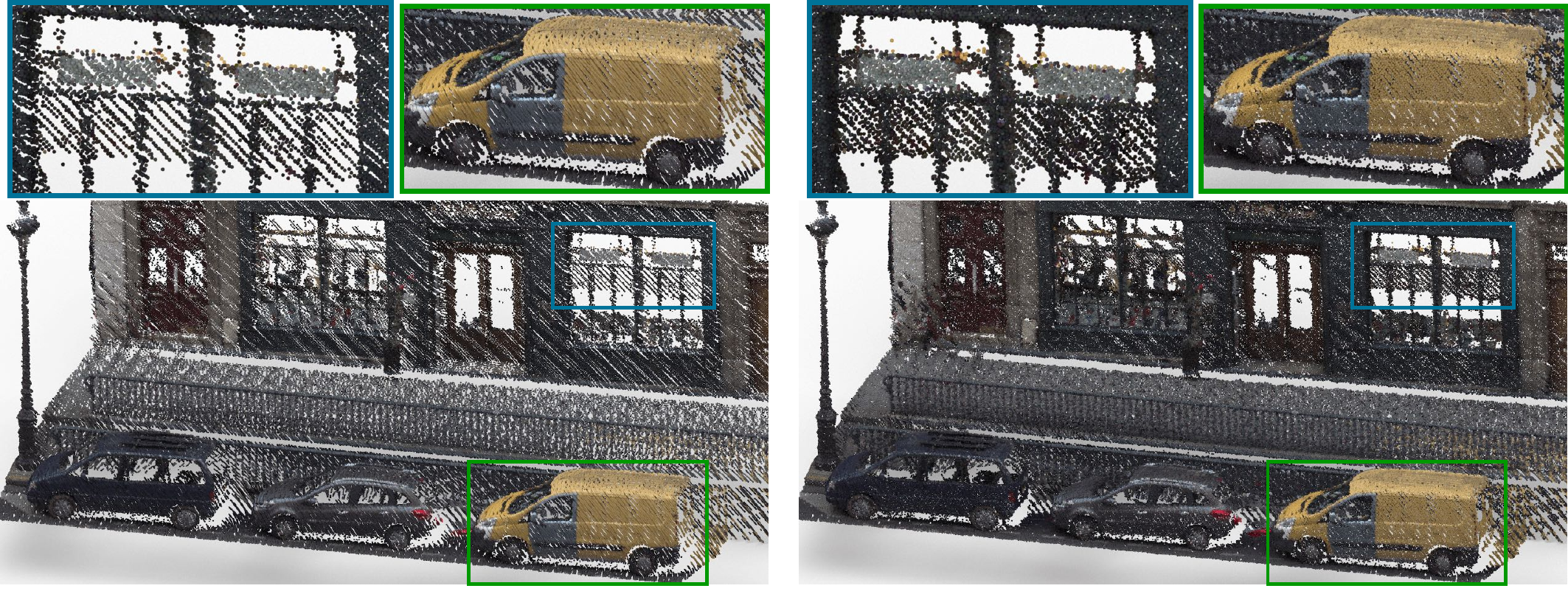}
\begin{tabularx}{\textwidth}{@{}Y@{}Y@{}}
(a) Noisy & (b) Ours \\
\end{tabularx}
\caption{Denoised results on Paris-CARLA-3D~\cite{deschaud2021pariscarla3d} dataset.}
\label{fig:supple-visual-Paris-CARLA-3D}
\end{figure}

\clearpage

\end{document}